\documentclass[sigconf]{acmart}

\copyrightyear{2022}
\acmYear{2022}
\setcopyright{acmcopyright}
\acmConference[MM '22]{Proceedings of the 30th ACM International Conference on Multimedia}{October 10--14, 2022}{Lisboa, Portugal}
\acmBooktitle{Proceedings of the 30th ACM International Conference on Multimedia (MM '22), October 10--14, 2022, Lisboa, Portugal}
\acmPrice{15.00}
\acmDOI{10.1145/3503161.3548088}
\acmISBN{978-1-4503-9203-7/22/10}

\AtBeginDocument{%
  \providecommand\BibTeX{{%
    \normalfont B\kern-0.5em{\scshape i\kern-0.25em b}\kern-0.8em\TeX}}}

\acmSubmissionID{1407}

\usepackage{bm}

\usepackage{amsthm, amsmath, amssymb}
\usepackage{mathrsfs}
\usepackage{multirow}
\usepackage{dsfont}
\usepackage[ruled,linesnumbered]{algorithm2e}
\usepackage{color}
\usepackage{enumitem}
\usepackage{hyperref}

\hypersetup{hidelinks,
            colorlinks=true,
            allcolors=black,
            pdfstartview=Fit,
            breaklinks=true}

\definecolor{CommentGreen}{RGB}{50,100,40}

\begin{document}

\title{DoF-NeRF: Depth-of-Field Meets Neural Radiance Fields }

\author{Zijin Wu}
\email{zijinwu@hust.edu.cn}
\affiliation{
  \institution{School of AIA, Huazhong University of Science and Technology}
}

\author{Xingyi Li}
\email{xingyi_li@hust.edu.cn}
\affiliation{
   \institution{School of AIA, Huazhong University of Science and Technology }
}

\author{Juewen Peng}
\email{juewenpeng@hust.edu.cn}
\affiliation{
  \institution{School of AIA, Huazhong University of Science and Technology}
}

\author{Hao Lu}
\email{hlu@hust.edu.cn}
\affiliation{
  \institution{School of AIA, Huazhong University of Science and Technology}
}

\author{Zhiguo Cao}
\email{zgcao@hust.edu.cn}
\authornote{Corresponding author.}
\affiliation{
   \institution{School of AIA, Huazhong University of Science and Technology}
}

\author{Weicai Zhong}
\email{zhongweicai@huawei.com}
\affiliation{
  \institution{CBG Search \& Maps BU, Huawei}
}

\renewcommand{\shortauthors}{Zijin Wu et al.}


\begin{abstract}
Neural Radiance Field (NeRF) and its variants have exhibited great success on representing 3D scenes and synthesizing photo-realistic novel views. However, they are generally based on the pinhole camera model and 
assume all-in-focus inputs. This limits their applicability as images captured from the real world often have finite depth-of-field (DoF). To mitigate this issue, we introduce DoF-NeRF, a novel neural rendering approach that can deal with shallow DoF inputs and can simulate DoF effect. In particular, it extends NeRF to simulate the aperture of lens following the principles of geometric optics. Such a physical guarantee allows DoF-NeRF to operate views with different focus configurations. Benefiting from explicit aperture modeling, DoF-NeRF also enables direct manipulation of DoF effect by adjusting virtual aperture and focus parameters. It is plug-and-play and can be inserted into NeRF-based frameworks. Experiments on synthetic and real-world datasets show that, DoF-NeRF not only performs comparably with NeRF in the all-in-focus setting, but also can synthesize all-in-focus novel views conditioned on shallow DoF inputs. An interesting application of DoF-NeRF to DoF rendering is also demonstrated. The source code will be made available at \href{https://github.com/zijinwuzijin/DoF-NeRF}{https://github.com/zijinwuzijin/DoF-NeRF}.
\end{abstract}



\maketitle

\begin{figure*}[t]
  \centering
  \setlength{\abovecaptionskip}{2pt}
  \includegraphics[width=\linewidth]{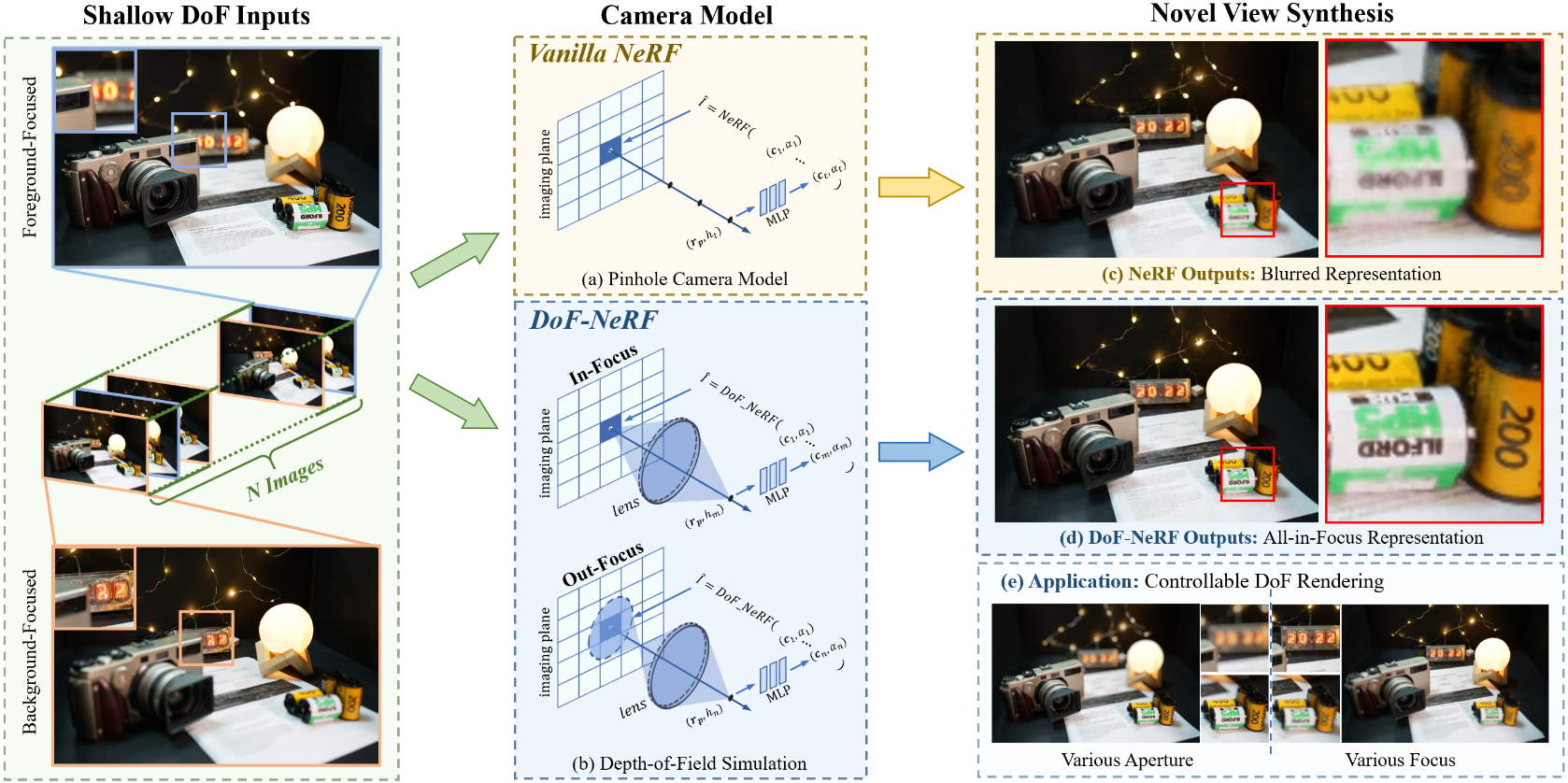}
  \caption{
  Given a set of shallow depth-of-field (DoF) images, (c) performance of NeRF deteriorates due to its (a) pinhole camera model assumption. We present DoF-NeRF, a novel NeRF-based framework using (b) explicit DoF simulation to represent (d) all-in-focus scenes from shallow DoF inputs. An application of DoF-NeRF to (e) DoF rendering is also presented. }
  \setlength{\belowcaptionskip}{2pt}
  \Description{Fig 1}
  \label{figure1}
\end{figure*}

\section{Introduction}
Novel view synthesis~\cite{debevec1996modeling, gortler1996lumigraph, levoy1996light} is a long-standing problem in computer vision and graphics. As synthesizing novel views of a 3D scene from a sparse set of input images is a fundamental task for applications in augmented reality (AR) and virtual reality (VR), 
a substantial amount of work has been 
conducted to seek for solutions. 
Recently Neural Radiance Field (NeRF)~\cite{mildenhall2020nerf}, an implicit multi-layer perceptron (MLP) based model that regresses colors and densities from 3D coordinates and 2D viewing directions, shows an impressive level of fidelity on novel view synthesis. In particular, NeRF 
uses classical volume rendering techniques~\cite{kajiya1984ray} to synthesize photo-realistic novel views by integrating the output colors and densities along emitted rays. Since the process of volume rendering is fully differentiable, NeRF can be optimized by minimizing the difference between the captured images and the rendered views.

However, NeRF and its variants~\cite{Martin-Brualla_2021_CVPR, kaizhang2020, deng2021depth} are generally based on the pinhole camera model and assume all-in-focus inputs, \textit{i.e.}, both foreground and background are clear. In reality, images captured from the real world often have finite depth-of-field (DoF). Namely, points of light that do not lie on the focal plane are imaged to a circular region on the sensor plane, rather than to single points. The size of the circular region (dubbed circle of confusion, or CoC) is affected by the diameter of the aperture (aperture size) 
and the distance from the camera to the focal plane (focus distance). Hence, photos captured with a small aperture usually present wide DoF, \textit{i.e.}, all objects are clear. 
In contrast, as the aperture 
diameter increases, objects that near the focal plane 
remain clear, but those far from the plane are blurred with a large CoC. This shallow DoF effect is ubiquitous in photography especially when shooting with a wide-aperture lens or taking close-up photos. 

While NeRF has shown astonishing results for novel view synthesis, 
its performance deteriorates when processing 
shallow DoF inputs. This can boil down to the assumption of the pinhole camera model---when predicting pixel colors, NeRF only 
considers the emission of spatial points on the ray passing through the pixel and neglects the scattered radiance from neighboring rays. To resolve this issue, we present DoF-NeRF, a novel NeRF-based framework that enables NeRF to tackle shallow DoF inputs. Specifically, 
we 
introduce three key changes: (1) a differentiable representation of CoC to simulate the radiance scattered between rays; (2) learnable parameters to enable direct manipulation of the 
DoF effect; (3) a patch-based ray selecting method for efficient optimization. Our key insight is to simulate the DoF effect by an optical-conforming radiance scattering method parameterized with two learnable parameters: aperture 
size and focus distance. The two parameters of each training view can guide the optimization of NeRF to generate a clear 3D scene representation. 
Our optical modeling of DoF 
can not only
synthesize all-in-focus novel views conditioned on shallow DoF inputs, but also provide highly controllable DoF rendering from novel 
viewpoints
. Our main contributions 
include the following.

\begin{itemize}[leftmargin=*]
\item We 
present DoF-NeRF, a novel neural rendering 
framework that can represent clear 3D scenes 
given shallow DoF inputs.
\item We introduce the Concentrate-and-Scatter technique, a plug-and-play rendering modification for NeRF-based methods to simulate the DoF effect.
\item We also contribute a new dataset for novel view synthesis of shallow DoF scenes. This dataset contains triplets of all-in-focus, foreground-focused, and background-focused 
from sparse viewpoints for each scene. Both real-world and synthetic data are included for further study.
\end{itemize}

\begin{figure*}[t]
  \centering
  \setlength{\abovecaptionskip}{2pt}
  \setlength{\belowcaptionskip}{2pt}
  \includegraphics[width=\linewidth]{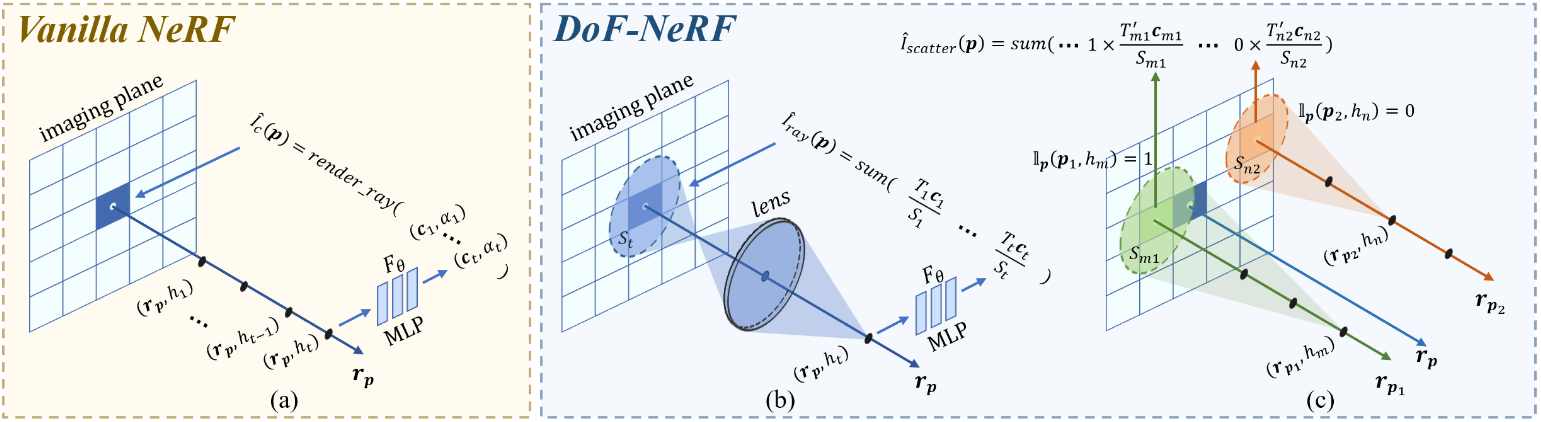}
  \caption{Basic principle of DoF-NeRF. The vanilla NeRF (a) adopts a pinhole camera model which ignores the DoF effect. DoF-NeRF 
  computes the observed color of ray $ \bm{r}_{\bm{p}} $ by adding (b) color diffusion $ \hat{I}_{ray}(\bm{p})$: the predicted radiance of spatial points on ray $\bm{r}_{\bm{p}}$ is diffused to a CoC on the imaging plane, 
  and (c) radiance scattering $ \hat{I}_{scatter}(\bm{p}) $: the scattered radiance of spatial points on neighboring rays, \textit{e.g.}, $ \bm{r}_{\bm{p}_1}, \bm{r}_{\bm{p}_2} $ may 
  affect the observed color of $\bm{r}_{\bm{p}}$. An indicator function is used to distinguish points by the CoC diameter and the distance to ray $\bm{r}_{\bm{p}}$. The observed color of ray $\bm{r}_{\bm{p}}$ is the combination of $ \hat{I}_{ray}(\bm{p}) $ and $ \hat{I}_{scatter}(\bm{p}) $. }
  \Description{Illus. of DoF-NeRF pipeline}
  \label{pipline}
\end{figure*}

\section{Related Work}

\subsection{Novel View Synthesis}

Novel view synthesis is a task of synthesizing novel camera perspectives from a set of input views and their corresponding camera poses. The research into novel view synthesis has a long history in computer vision and graphics community. 
Various approaches are investigated, including image-based rendering~\cite{debevec1996modeling, gortler1996lumigraph, levoy1996light} and explicit geometric representations such as voxel grids~\cite{liao2018deep,jimenez2016unsupervised,xie2019pix2vox}, point clouds~\cite{achlioptas2018learning,fan2017point}, triangle meshes~\cite{kanazawa2018learning,ranjan2018generating,wang2018pixel2mesh}, and multiplane images (MPIs)~\cite{zhou2018stereo,flynn2019deepview,tucker2020single}. Recent studies~\cite{NEURIPS2019_SRN, mildenhall2019local, sitzmann2019deepvoxels, 2019Neural} have shown the superiority of implicit representations in rendering high quality novel views. For example, Mildenhall~\textit{et al.} propose Neural Radiance Field (NeRF)~\cite{mildenhall2020nerf}, an implicit MLP-based model that maps 3D coordinates plus 2D viewing directions to opacity and color values, which is capable of representing a complex 3D scene and rendering photo-realistic novel views. 

However, 
drawbacks of NeRF remain, including entailing immense posed images and its high computational requirements for rendering novel views. To mitigate these issues, researchers have introduced improvements upon NeRF to extend its performance and applicability, such as faster training~\cite{deng2021depth, sun2021direct}, faster inference~\cite{reiser2021kilonerf, Garbin21arxiv_FastNeRF}, optimizing NeRF with low-light~\cite{mildenhall2021nerf} or high dynamic range~\cite{huang2021hdr} images, 
improving generalization~\cite{Schwarz20neurips_graf, niemeyer2021giraffe}, and representing dynamic scenes~\cite{li2020neural, xian2021space}. Recently, 
Self-Calibrating NeRF~\cite{jeong2021self} combines the pinhole camera, radial distortion, and a generic non-linear camera distortion for self-calibration by modeling distortion parameters. However, all these methods neglect the DoF effect. In this paper, we extend NeRF to simulate the aperture of lens and model the DoF effect by optimizing two learnable parameters, \textit{i.e.}, aperture size 
and focus distance. 
This enables the synthesis of all-in-focus novel views with shallow DoF inputs as well as the rendering of 3D scenes with arbitrary aperture and focus distance settings.

\subsection{DoF Rendering}

Rendering DoF effects from a single all-in-focus image has been well studied in previous work. Some work~\cite{dutta2021stacked,ignatov2019aim,ignatov2020aim,ignatov2020rendering,qian2020bggan} directly regresses a shallow DoF image using neural networks. However, these methods cannot adjust DoF effects as they are trained on EBB! dataset~\cite{ignatov2020rendering} which only provides pairs of wide and shallow DoF images. The DoF effects in \cite{busam2019sterefo,peng2021interactive,wadhwa2018synthetic,xian2021ranking,xiao2018deepfocus,zhang2019synthetic} are controllable but usually require an extra disparity map. Although the disparity map can be predicted by depth estimation, 
its accuracy is not guaranteed. Another challenge is the revealing of invisible background objects during 
rendering, because no image of other views is provided. 

To address these problems, several methods focusing on synthesizing DoF effects on NeRF models have been proposed recently. RawNeRF~\cite{mildenhall2021nerf} adopts a multi-plane representation to render DoF effects, while NeRFocus~\cite{wang2022nerfocus} proposes a frustum-based volume rendering to approximate the imaging of a thin lens model. However, both studies mentioned above still assume all-in-focus inputs. In this work, through 
appropriate optical modeling, we optimize the aperture diameters and the focus distances during training and 
render clear novel views with shallow DoF inputs. 

\section{Preliminary}
In this section, we briefly review 
the principle of NeRF.
To represent a scene, NeRF optimizes a continuous function parameterized by an MLP network $ G_\Theta:(\bm{x},\bm{d})\rightarrow(\bm{c},\alpha) $ 
which maps
a spatial position $ \bm{x}=(x,y,z) $ and viewing direction $ \bm{d}=(\theta,\varphi)  $ to its corresponding emitted color $ \bm{c} $ and transparency $ \alpha $. 
the observed color $ \hat{I}_n(\boldsymbol{p}) $ of the pixel $ \boldsymbol{p} $ can be represented as an integral of all 
emitted colors weighted by the 
opaqueness
along the camera ray $ \bm{r}_{\bm{p}}(h) $, or can be written as the weighted radiance at $N_s$ sample points along a ray:
\begin{equation}
    \hat{I}_n(\bm{p})=\sum_{i=1}^{N_{s}}T_i(1-\alpha(\bm{r}_{\bm{p}}(h_i),\Delta_i))\bm{c}(\bm{r}_{\bm{p}}(h_i), \bm{d})\,,
\end{equation}
where $ \Delta_i=h_{i+1}-h_i$ and 
\begin{equation}
    T_i=\prod_{j=1}^{i-1}\alpha(\bm{r}_{\bm{p}}(h_j),\Delta_j)
\end{equation}
denotes the accumulated transmittance along the ray, \textit{i.e.}, the probability of the emitted light travelling from $ h_1 $ to $ h_i $ without hitting other particles. 
Computing the color value of each pixel on the imaging plane via volume rendering above composites a complete image.

To optimize
the continuous function $G_\Theta$ 
from a set of input images $ \mathcal{I} = \{I_1,I_2,...I_{N}\}$, 
NeRF adopts the photometric error between the synthesized views $ \hat{\mathcal{I}} $ and corresponding 
observations
$ \mathcal{I} $:
\begin{equation}
    \mathcal{L}=\sum_{n=1}^{N}\Vert I_n-\hat{I}_n\Vert_2^2\,.
\end{equation}

Note that, vanilla NeRF 
assumes a linear pinhole camera model
where the color value of 
a 
pixel $ \boldsymbol{p}=(u,v) $ on the imaging plane is only determined by 
a single camera ray $ \boldsymbol{r}(h)=\bm{o}+h\bm{d} $ that travels from the camera center $ \bm{o} $ and passes through the pixel $ \boldsymbol{p} $ along the viewing direction $ \bm{d} $. Therefore, NeRF ignores the interference between rays and does not model the aperture structure, 
leading to its 
incapability to simulate the DoF effect.

\section{Approach}
Instead of 
assuming
the linear pinhole camera model in volume rendering, 
we extend NeRF to 
incorporate
the representation of CoC to simulate the DoF. 
This
allows NeRF to 
approximate the DoF effect 
under arbitrary aperture size and focus distance configuration 
such that all-in-focus scenes can be represented with shallow DoF 
inputs.

The emergence of DoF 
consists of two steps:
radiance of the spatial points scattering to a particular area and center ray accepting the scattered radiance from neighboring rays. We 
implement
these two steps by combining optical models and classical volume rendering (Fig.~\ref{pipline}). To optimize NeRF with shallow DoF images, 
we 
introduce
two learnable parameters: the aperture parameters 
$\mathcal{K}=\{K_1,K_2,...K_N\}$ and the focus distances $ \mathcal{F}=\{F_1,F_2,...F_N\}$ for $ N $ images in the training set $\mathcal{I}$. The aperture parameters and focus distances are jointly optimized 
in the training process. 

In what follows, we first introduce our explicit aperture modeling and then 
explain how DoF-NeRF is optimized. The table of notations 
can be found in the appendix.

\subsection{DoF in Radiance Field}
The physical model of imaging and DoF has been well studied in geometric optics~\cite{hecht2017optics, li2005applied}.
Assume that we neglect the ray distortions 
caused 
by the lens. In an ideal optical system, a spatial point $ \bm{p} $ with the point-to-lens distance (object distance) $ h_t $ is projected to a circular region (CoC) on the imaging plane. 
The diameter of the region $ \delta(h_t) $ can be determined by the focal length $ f $, aperture diameter $ D $, and focus distance $ F $, which 
amounts to the following equation
\begin{equation}
    \delta(h_t) =  fD\times\frac{\left|h_t-F\right|}{h_t(F-f)}\,.
    \label{eq:delta-t}
\end{equation}

Since the focus distance $ F $ and object distance $ h_t $ are often 
much
larger than the focal length $ f $, we can 
modify 
Eq.~\eqref{eq:delta-t} 
such that
\begin{equation}
    \delta(h_t) = fD\times\frac{\left| h_t-F\right|}{Fh_t}=fD\times\left|\frac{1}{F}-\frac{1}{h_t}\right| = K \times\left|\frac{1}{F}-\frac{1}{h_t}\right| \,,
\end{equation}
where the product of focal length and aperture diameter can be replaced by an aperture parameter $K=f\times D$.

For simplicity, we assume that the emitted radiance $ \hat{C}(\boldsymbol{p}, h_t) $ of the spatial point evenly disperses to a CoC of diameter $ \delta (h_t) $:
\begin{equation}
    \hat{C}(\boldsymbol{p}, h_t)=\frac{\left(1-\alpha\left(\bm{r}_{\bm{p}}(h_t), \Delta_t\right)\right)\bm{c}\left(\bm{r}_{\bm{p}}(h_t), \bm{d}\right)}{S(h_t)}\,,
    \label{eq:C}
\end{equation}
where $ S(h_t) = \pi \delta^2 (h_t) /4 $ denotes the area of the CoC.

Different from the pinhole camera assumption, the observed color of the ray $ \bm{r}_{\bm{p}} $ can 
also 
be affected by spatial points whose CoC radius is larger than the distance to the ray $ \bm{r}_{\bm{p}} $. 
Apparently, points on neighboring rays only contribute to the radiance accumulation without affecting 
the transmittance $ T(\bm{r}_{\bm{p}}) $. 
Given $M$ rays specific to pixels $ \bm{\mathcal{P}}=\{\bm{p}_1,\bm{p}_2,...,\bm{p}_M\} $ on which the scattered radiance of spatial points may affect the color prediction of center ray $\bm{r}_{\bm{p}}$, 
we can 
compute the color $ \hat{I} $ of pixel $ \bm{p} $ by 
summing up 
emitted radiance.  
without changing the transmittance coefficient. 
Specifically, computing the observed color $\hat{I}(\bm{p}) $  can be split into two parts: $ \hat{I}_{ray}(\bm{p}) $ and $ \hat{I}_{scatter}(\bm{p}) $ which respectively represent the diffused radiance on ray $ \bm{r}_{\bm{p}} $ and the scattered radiance from other rays. 
This takes the form
\begin{gather}
    \hat{I}_{ray}(\bm{p}) = \sum_{i=1}^{N_s}T_i\hat{C}(\boldsymbol{p}, h_i)\,,\\
    \hat{I}_{scatter}(\bm{p})=\sum_{j=1}^{M}\sum_{i=1}^{N_s}\mathds{1}_{\delta(h'_i)>2\Vert\bm{p}-\bm{p}_j\Vert_2^2}T'_{ij}\hat{C}(\boldsymbol{p}_j, h'_i)\,,\\
    \hat{I}(\bm{p})=\hat{I}_{ray}(\bm{p})+\hat{I}_{scatter}(\bm{p})\,,
    \label{eq:Ip}
\end{gather}
where $h_i$ and $h'_i$ denote the depth of the $i$-th sample point on the center ray and neighboring rays, respectively; $ T_i $ denotes the accumulated transmittance of the $i$-th sample point on the ray $ \bm{r}_{\bm{p}} $ 
, and 
 $ T'_{ij} $ denotes the transmittance of scattered radiance from the $i$-th sample point on the ray specific to pixel $\bm{p}_j$. 
 An indicator function $\mathds{1}$ is used to distinguish points by the CoC diameter $ \delta(h'_i) $ and the distance to ray $\bm{r}_{\bm{p}}$.

\begin{figure}[t]
  \centering
  \setlength{\abovecaptionskip}{2pt}
  \setlength{\belowcaptionskip}{2pt}
  \includegraphics[width=\linewidth]{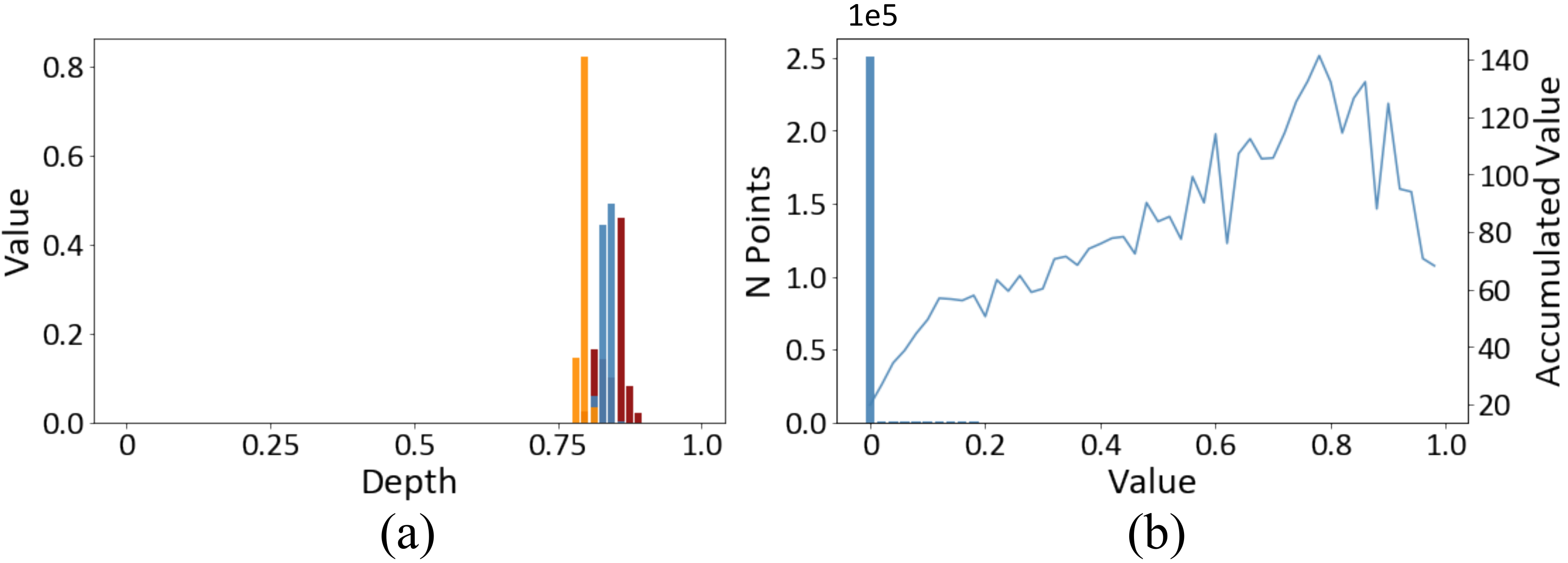}
  \caption{ Visualization of volume coefficients $ K_{volume} $. In (a), we randomly choose $3$ rays from the 3D scene and visualize the volume coefficient $ K_{volume} $ of sample points on each ray according to the depth. 
  Different colors in (a) denote 
  different rays. In (b) the histogram shows the value distribution of $ K_{volume} $ of $1024$ selected rays, and the polyline shows the accumulated $ K_{volume} $ of points in each bin of the histogram.  }
  \Description{vis. of K_volume}
  \label{kvolume}
\end{figure}

However, 
computing the color of the ray $ \bm{r}_{\bm{p}} $ directly following Eq.~\eqref{eq:C}$\sim$\eqref{eq:Ip} can be rather inefficient. It requires traversing all the spatial points on neighboring rays to 
obtain the 
color of the center rays. 

To mitigate this issue, we analyze a fact of the volume rendering and propose our solution in what follows.
For spatial points on the ray $ \bm{r}_{\bm{p}} $, 
we evaluate their contributions 
to the color prediction 
with a volume rendering coefficient $ K_{volume} $:
\begin{equation}
    K_{volume}(\bm{r}_{\bm{p}}, h_t)=T_t\left(1-\alpha\left(\bm{r}_{\bm{p}}(h_t), \Delta_t\right)\right)\,.
\end{equation}
We randomly choose $1024$ rays from input views and visualize the distribution of the volume coefficients (see Fig.~\ref{kvolume}). For most 
rays, points with large volume coefficients gather in a narrow depth interval. The distribution of the volume coefficient indicates that most 
spatial points 
do not affect the observed color but 
incur 
considerable computational overhead. 

\begin{figure}[t]
  \centering
  \setlength{\abovecaptionskip}{2pt}
  \setlength{\belowcaptionskip}{2pt}
  \includegraphics[width=\linewidth]{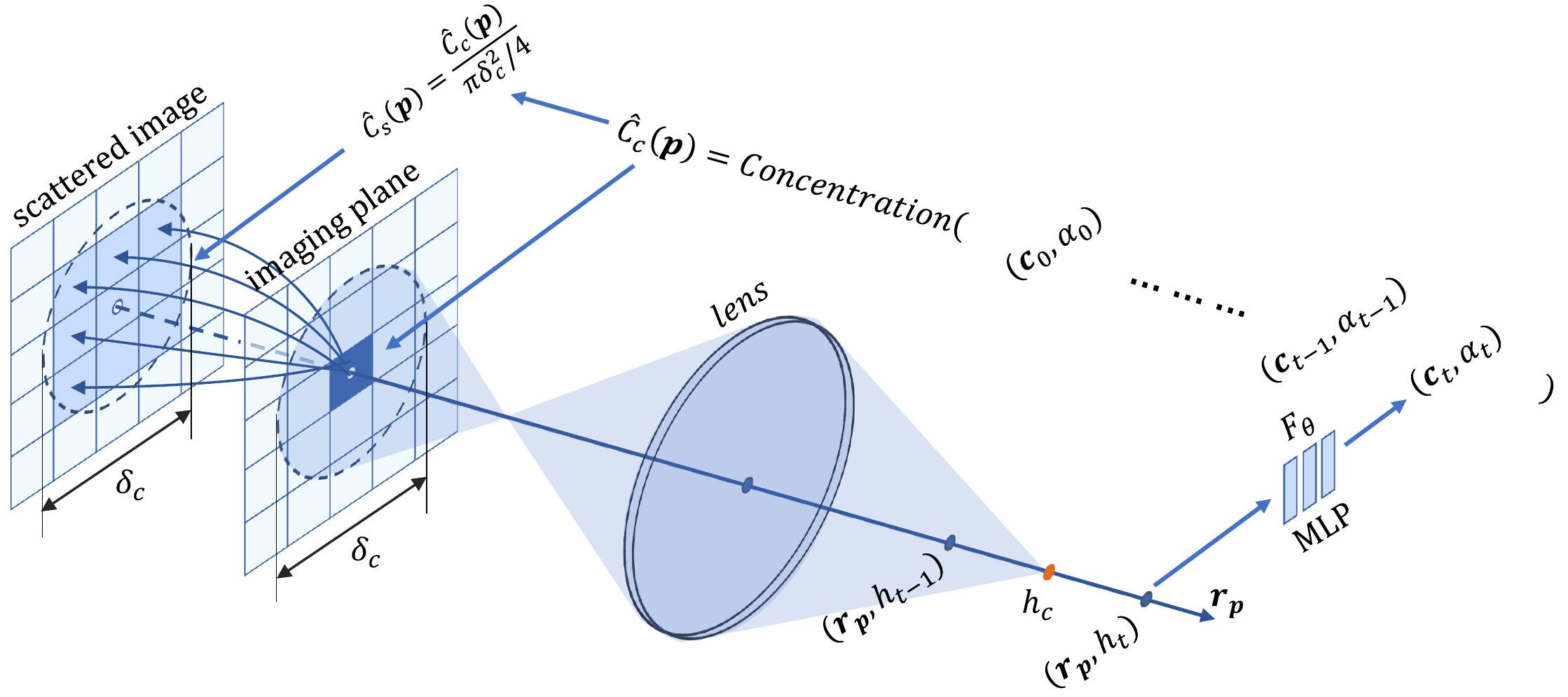}
  \caption{ Illustration of our concentrate-and-scatter rendering. To reduce the cost of computation, we first concentrate the emitted radiance of points on the ray $ \bm{r}_{\bm{p}} $ (right side of the imaging plane) to the concentration depth $h_c$ (orange point) and calculate the corresponding CoC diameter. Then we scatter the concentrated radiance to 
  the
  neighboring areas 
  within its CoC diameter
  (left side of the imaging plane). }
  \Description{Illus. of con-n-scat method}
  \label{connscat}
\end{figure}

Nonetheless,
simply ignoring those points may lead to the brightness change of 
rendered 
images. 
In the view of 
the volume coefficients distribution, we introduce the idea of concentrate-and-scatter rendering. As shown in Fig.~\ref{connscat}, the core idea 
is to concentrate the emitted radiance of all the spatial points along the ray to the concentration depth $ h_c $, and to scatter the concentrated radiance to its corresponding CoC.
Since the scattered radiance does not affect the 
transmittance on the ray $\bm{r}_{\bm{p}}$, the concentration of the radiance follows the original volume rendering. For $N_s$ sampled points along the ray $ \bm{r}_{\bm{p}} $, the concentration depth $ h_c $, CoC diameter $ \delta_c $ of the concentration point, and the scattered radiance $ \hat{C}_s(\boldsymbol{p}) $ can be formulated by
\begin{gather}
    h_c(\bm{p}) = \frac{\sum_{i=1}^{N_s}K_{volume}(\bm{r}_{\bm{p}}, h_t)h_i}{\sum_{i=1}^{N_s}K_{volume}(\bm{r}_{\bm{p}}, h_t)} \,,\\
    \delta_c(\bm{p}) = K\left|\frac{1}{F}-\frac{1}{h_c}\right| \,,\\
    \hat{C}_s(\boldsymbol{p})=\frac{\sum_{i=1}^{N_s}K_{volume}(\bm{r}_{\bm{p}}, h_t)\bm{c}(\bm{r}_{\bm{p}}(h_i), \bm{d})}{\pi\delta_c^2(\bm{p})/4}\,.
\end{gather}
For $P$ rays 
with respect to 
the pixels $ \bm{\mathcal{P}}=\{\bm{p}_1,\bm{p}_2,...\bm{p}_P\} $ whose radiance may 
affect the color prediction of ray $\bm{r}_{\bm{p}}$, including the ray $\bm{r}_{\bm{p}}$ itself, the observed color of ray $\bm{r}_{\bm{p}}$ 
can be defined by
\begin{equation}
    \hat{I}(\bm{p})=\sum_{j=1}^{P}\mathds{1}_{\delta_c(\bm{p})>2\Vert\bm{p}-\bm{p}_j\Vert_2^2}\hat{C}_s(\bm{p}_j)\,.
\end{equation}

We implement the algorithm with the \texttt{CuPy} package to achieve a significant speedup. The detail of the 
algorithm can be found in the appendix.

\subsection{Ray Selection}

When optimizing the scene representation, vanilla NeRF randomly chooses $ N_{rand} $ rays from all input views. 
Since the simulation of the DoF effect 
requires to consider
radiance 
scattered
from neighboring rays, 
a straight-forward approach is to 
compute several rays 
that may affect the center ray. This method 
is of low efficiency
in training as it only considers the observed color of the center ray in a single iteration. 
Yet
another approach is to 
compute the concentrated radiance of all rays from one imaging plane and to scatter every pixels on the whole imaging plane. However, it 
entails to compute
every ray of the whole imaging plane in each iteration, which leads to an unacceptable cost in computing.

\begin{figure}[t]
  \centering
  \setlength{\abovecaptionskip}{2pt}
  \setlength{\belowcaptionskip}{2pt}
  \includegraphics[width=\linewidth]{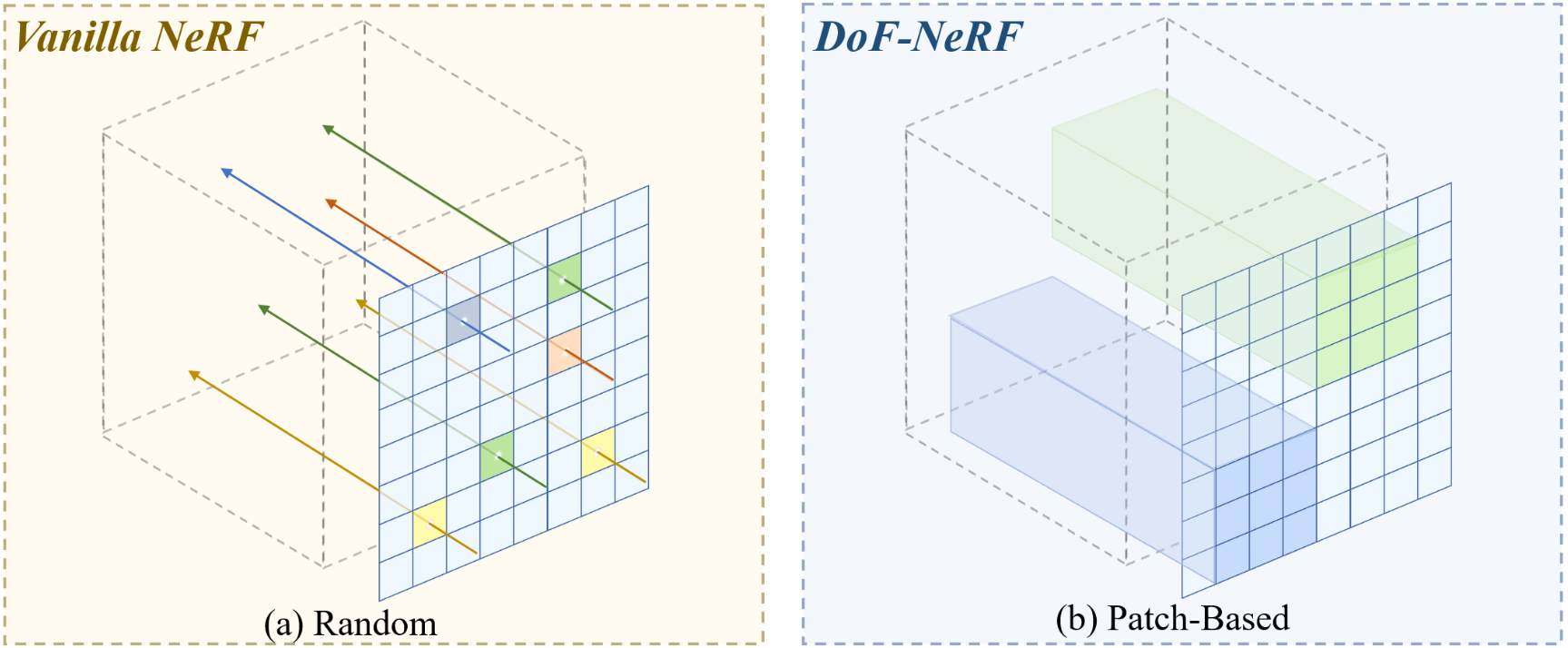}
  \caption{ Illustration of our patch-based optimization. 
Instead of 
randomly choosing rays (a) as in NeRF, we optimize 
DoF-NeRF by selecting rays from a concentrated area in each iteration (b).
  }
  \Description{Illus. of con-n-scat method}
  \label{patchify}
\end{figure}

We adopt a patch-based ray selection method (Fig.~\ref{patchify}) which can be considered as a compromise between the two approaches above. 
We construct a group of anchors, where each anchor is set every $N_{anchor}$ pixels
on 
the
imaging plane of inputs
and 
determines the center of an $N_{patch}\times N_{patch}$ patch. In each iteration, we randomly choose one patch and 
guide rays passing though pixels in the selected patch. Observed colors of the pixels are computed using the concentrate-and-scatter rendering mentioned above.

\begin{algorithm}[t]
    \small
    \caption{Joint Optimization of DoF-NeRF}
    \label{alg1}
    \KwIn{$ N $ images $ \mathcal{I} = \{I\}_{i=1}^{N} $}
    \KwOut{NeRF Model $ G_\Theta $, 
	       aperture $ [\hat{K}]_{i=1}^{N} $, 
	       focus distance $[\hat{F}]_{i=1}^{N}$}
	import torch.nn as nn \\
	$ [\hat{K}] $ = nn.Parameter(shape=(N, 1), requires\_grad=True) \\
	$ [\hat{F}] $ = nn.Parameter(shape=(N, 1), requires\_grad=True) \\
	$ G_\Theta $ = NeRF\_Model(requires\_grad=True) \\
	\For{i in range($N_{iters}$)}
	{
	    \eIf{$i<N_{pretrain}$}
	    {
	        $[\bm{r}]_i, I_i$=random\_rays($ \mathcal{I} $) \\
	        $\hat{I}_i$=Volume\_Rendering($ G_\Theta, [\bm{r}]_i $)  \textcolor{CommentGreen}{\# Sec. 3} \\
	        $\mathcal{L}$=loss($\hat{I}_i, I_i$)  \textcolor{CommentGreen}{\# Eq. 3} \\
	        $\mathcal{L}$.backward( ) \\
	        optimizer.update($\hat{\Theta}$) \\
	    }{
	        $[\bm{r}]_i, I_i$=patch\_rays($ \mathcal{I} $) \textcolor{CommentGreen}{\# Sec. 4.2} \\
	        $ [\hat{\bm{C}}]_i, [h_{c}]_i$=Concentration($ G_\Theta, [\bm{r}]_i $) \textcolor{CommentGreen}{\# Eq. 13} \\
	        $[\delta_c]_i$=CoC\_Radius($[h_c], \hat{K}_i,\hat{F}_{i}$) \textcolor{CommentGreen}{\# Eq. 12} \\
	        $\hat{I}_i$=Scatter($[\hat{\bm{C}}]_i, [\delta_c]_i$)  \textcolor{CommentGreen}{\# Eq. 14} \\
	        $\mathcal{L}$=loss($\hat{I}_i, I_i$)  \textcolor{CommentGreen}{\# Eq. 3} \\
	        $\mathcal{L}$.backward( ) \\
	        optimizer.update($\hat{\Theta}, [\hat{K}], [\hat{F}]$) \\
	    }
	}
	
\end{algorithm}

\subsection{Joint Optimization}
Although
using randomly chosen rays in the simulation of DoF effect is unpractical, it shows high efficiency in optimizing the geometric representation of 3D scenes. The patch-based method, however, often leads to divergence or sub-optimal results due to the gathering of rays. Thus, we 
resort to
a two-stage optimization process to reduce the complexity of learning geometry representation, aperture size, and focus distance. 

At the first stage, we train the NeRF network with the aperture parameter set to $0$, which degenerates the rendering model to a linear pinhole camera. In this stage, the aperture parameters and focus distances are not 
optimized, and the classical ray-choosing and volume rendering 
are adopted. This stage aims to generate a coarse 3D representation where the rendered foreground and background may be blurred due to the shallow DoF inputs. 
At
the
second stage, 
we further optimize the NeRF network with the concentrate-and-scatter method using patch-based ray selection. The aperture parameters, focus distances, and the NeRF network parameters are jointly optimized. 
We summarize our 
learning algorithm in Algorithm~\ref{alg1}.

\begin{figure*}[t]
  \setlength{\abovecaptionskip}{2pt}
  \setlength{\belowcaptionskip}{2pt}
  \centering
  \includegraphics[width=\linewidth]{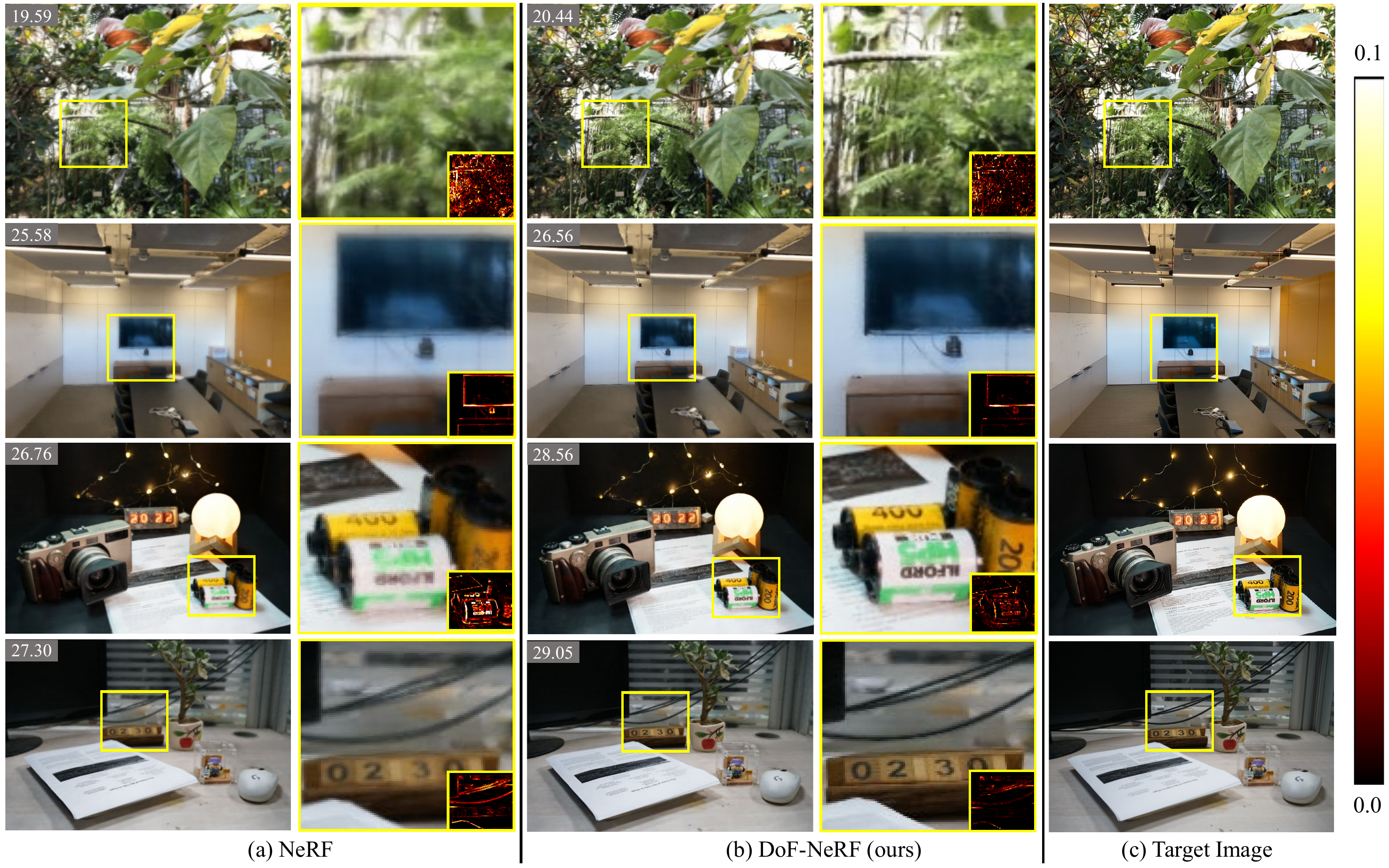} 
  \caption{Comparison of NeRF and our approach in all-in-focus rendering with shallow DoF inputs. The first and third column shows the images rendered by NeRF and our approach, respectively. PSNR is shown at the upper left corner. 
  We zoom in all the yellow boxes 
  in the second and fourth column with error maps ($0$ to $0.1$ pixel intensity range) shown at the lower right corner. The first two rows show the result of scenes from the synthetic dataset, and the third and fourth rows are scenes from the real-world dataset.
  }
  \Description{Visualization of main exp in synthetic dataset.}
  \label{mainexp}
\end{figure*}

\begin{table}[t]
  \caption{Comparison of NeRF~\cite{mildenhall2020nerf} and our framework in the real-world dataset.}
  \label{tab:exp-rw}
  \begin{tabular}{ccccl}
    \toprule
    Scene & Model & PSNR$ (\uparrow) $ & SSIM$ (\uparrow) $ & LPIPS$ (\downarrow) $\\
    \midrule
    \multirow{2}*{amiya} & NeRF & 26.924 & 0.9092 & 0.1633 \\
    ~ & ours & $ \textbf{28.311} $ & $ \textbf{0.9289} $ & $ \textbf{0.1370} $ \\
    \multirow{2}*{camera} & NeRF & 25.593 & 0.8862 & 0.1574 \\
    ~ & ours & $ \textbf{27.714} $ & $ \textbf{0.9134} $ & $ \textbf{0.1259} $ \\
    \multirow{2}*{plant} & NeRF & 28.272 & 0.8961 & 0.1581 \\
    ~ & ours & $ \textbf{30.317} $ & $ \textbf{0.9290} $ & $ \textbf{0.1178} $\\
    \multirow{2}*{turtle} & NeRF & 33.531 & 0.9566 & 0.0939  \\
    ~ & ours & $ \textbf{34.965} $ & $ \textbf{0.9647} $ & $ \textbf{0.0823} $\\
  \bottomrule
\end{tabular}
\end{table}

\section{
Results and Discussions
}

\subsection{Dataset and Evaluation}
We evaluate our method on both a real-world dataset and a synthetic dataset. 
The real-world dataset consists 
of
$7$ scenes, where each contains $20\sim30$ image triplets. Each triplet includes an all-in-focus image taken with small aperture
and two images taken with large aperture focusing on the foreground and background, respectively. 
We generate camera parameters by COLMAP~\cite{schonberger2016structure} using all-in-focus images.
The synthetic dataset is generated based on depth estimation and a recent single-image DoF rendering framework: for each
image in the Real Forward-Facing dataset~\cite{mildenhall2020nerf}, we use DPT~\cite{Ranftl_2021_ICCV} to generate the disparity map and 
BokehMe~\cite{Peng2022BokehMe} to render 
shallow DoF images. 
Details of the datasets can be found in the appendix.

\begin{table}[t]
  \caption{Comparison of NeRF~\cite{mildenhall2020nerf} and our framework in the synthetic dataset.}
  \label{tab:exp-syn}
  \begin{tabular}{ccccl}
    \toprule
    Scene & Model & PSNR $ (\uparrow) $ & SSIM$ (\uparrow) $ & LPIPS$ (\downarrow) $\\
    \midrule
    \multirow{2}*{fortress} & NeRF & 28.142 & 0.7826 & 0.2011\\
    ~ & ours & $ \textbf{29.168} $ & $ \textbf{0.8099} $ & $ \textbf{0.1830} $ \\
    \multirow{2}*{leaves} & NeRF & 19.450 & 0.6541 & 0.3190 \\
    ~ & ours & $ \textbf{20.025} $ & $ \textbf{0.7000} $ & $ \textbf{0.2766} $ \\
    \multirow{2}*{room} & NeRF & 26.668 & 0.8743 & 0.1961 \\
    ~ & ours & $ \textbf{29.443} $ & $ \textbf{0.9135} $  & $ \textbf{0.1502} $ \\
    \multirow{2}*{trex} & NeRF & 24.433 & 0.8379 & 0.1723 \\
    ~ & ours & $\textbf{25.726}$ & $\textbf{0.8744}$ & $\textbf{0.1564}$\\
  \bottomrule
\end{tabular}
\end{table}

\begin{figure*}[t]
  \centering
  \setlength{\abovecaptionskip}{2pt}
  \setlength{\belowcaptionskip}{2pt}
  \includegraphics[width=\linewidth]{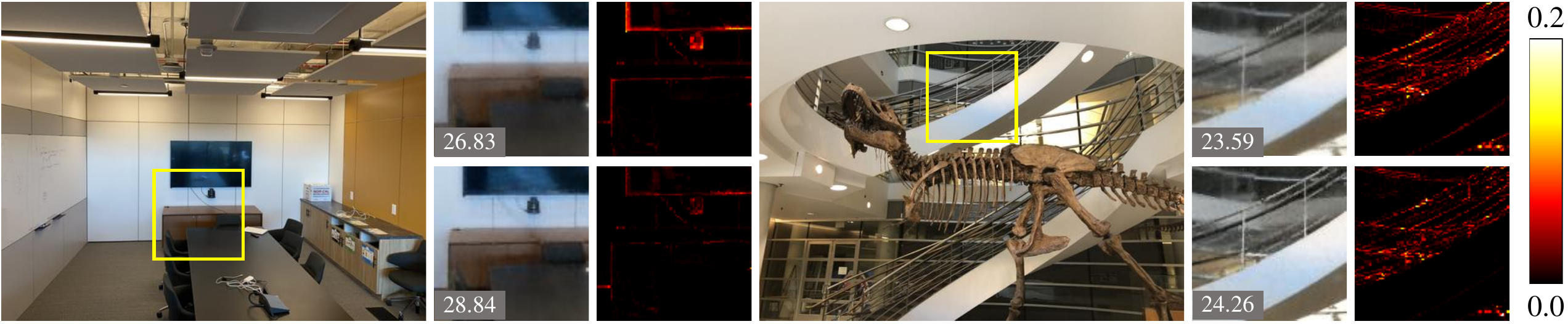} 
  \caption{ Comparison of DS-NeRF~\cite{deng2021depth} and our approach in the synthetic dataset. For each scene, the zoom-in of rendered images and error maps ($0$ to $0.2$ pixel intensity range) are presented. They are respectively obtained from DS-NeRF (first row) and our model (second row). For each subfigure, PSNR is shown on the lower left. }
  \Description{Visualization of main exp in synthetic dataset.}
  \label{dsnerf}
\end{figure*}

Following~\cite{mildenhall2020nerf}, we 
adopt 
PSNR, SSIM~\cite{wang2004image} and LPIPS~\cite{zhang2018unreasonable} as evaluation metrics.
In all the experiments, 
all images are of $ 497\times331 $ resolution for the real-world dataset and $ 504\times378 $ for the synthetic dataset. 
$1/9$ of the images are 
held
out for 
testing.
$50\%$ foreground-focused images and $50\%$ background-focused images make up the training images. 
Both our method and vanilla NeRF adopt the same mixing order of images in each scene.

\subsection{Improvement over Vanilla NeRF}
In this section, 
we compare DoF-NeRF qualitatively and quantitatively
with
vanilla NeRF on the real-world and synthetic datasets. 
We set the aperture parameters of the DoF-NeRF to zero 
and render all-in-focus novel views for evaluation. 
As shown in Table~\ref{tab:exp-rw} and Table~\ref{tab:exp-syn}, one can observe that our method demonstrates better perceptual qualities than vanilla NeRF when rendering all-in-focus novel views on both two datasets. 
This implies that our explicit aperture modeling enables NeRF to tackle shallow DoF inputs and benefits all-in-focus novel view synthesis.
We also provide the qualitative comparisons in Fig.~\ref{mainexp}. As can be seen, vanilla NeRF is prone to generate blurry renderings and miss some texture details. In contrast, our DoF-NeRF can synthesize photo-realistic all-in-focus novel views with fine-grained details.

\subsection{Improvement over DS-NeRF}

Since our DoF-NeRF is a modification 
of 
volume rendering, it can work on variants of NeRF as a plug-and-play module. 
To show this,
we substitute the NeRF architecture 
with DS-NeRF~\cite{deng2021depth} and 
use
the same optimization technique. We then compare DS-NeRF and our model in the synthetic dataset. As shown in Table~\ref{tab:exp-dsnerf}, 
the inclusion
of our DoF module results in better rendering 
quality with shallow DoF images as inputs. The qualitative results are visualized in Fig.~\ref{dsnerf}. 


\subsection{Comparison on All-in-Focus Inputs}
Here we present the comparison on the all-in-focus dataset. To 
additionally validate the effectiveness of our method in the all-in-focus setting, we design an experiment where wide DoF images from the Real Forward-Facing dataset~\cite{mildenhall2020nerf} are used as inputs and adopt the same initialization, training, and rendering settings as the experiments conducted on the shallow DoF data. Table~\ref{tab:exp-aif} reports the 
quantitative comparison on the all-in-focus dataset. Although our framework is originally designed for shallow DoF inputs, experiments indicate that it shows comparable performance against vanilla NeRF using all-in-focus images as inputs. 

\begin{table}[t]
  \setlength{\abovecaptionskip}{2pt}
  \setlength{\belowcaptionskip}{2pt}
  \caption{Comparison of DS-NeRF~\cite{deng2021depth} and our framework in the synthetic dataset.}
  \label{tab:exp-dsnerf}
  \begin{tabular}{ccccc}
    \toprule
    Scene & Model & PSNR$ (\uparrow) $ & SSIM$ (\uparrow) $ & LPIPS$ (\downarrow) $\\
    \midrule
    \multirow{2}*{fortress} & DS-NeRF & 28.493 & 0.7609 & 0.2373\\
    ~ & ours & $ \textbf{29.704} $ & $ \textbf{0.8112} $ & $ \textbf{0.1970} $ \\
    \multirow{2}*{leaves} & DS-NeRF & 17.872 & 0.5096 & 0.4358 \\
    ~ & ours & $ \textbf{18.312} $ & $ \textbf{0.5689} $ & $ \textbf{0.3840} $ \\
    \multirow{2}*{room} & DS-NeRF & 26.8301 & 0.8641 & 0.2287 \\
    ~ & ours & $ \textbf{29.550} $ & $ \textbf{0.9045} $  & $ \textbf{0.1817} $ \\
    \multirow{2}*{trex} & DS-NeRF & 22.334 & 0.7063 & 0.3438 \\
    ~ & ours & $\textbf{23.600}$ & $\textbf{0.7764}$ & $\textbf{0.2773}$\\
  \bottomrule
\end{tabular}
\end{table}

\begin{table}[t]
  \setlength{\abovecaptionskip}{2pt}
  \setlength{\belowcaptionskip}{2pt}
  \caption{Comparison of NeRF~\cite{mildenhall2020nerf} and our framework in the all-in-focus dataset.}
  \label{tab:exp-aif}
  \begin{tabular}{ccccc}
    \toprule
    Scene & Model & PSNR$ (\uparrow) $ & SSIM$ (\uparrow) $ & LPIPS$ (\downarrow) $\\
    \midrule
    \multirow{2}*{fortress} & NeRF & 33.973 & 0.9411 & 0.0469\\
    ~ & ours & $ \textbf{34.007} $ & $ \textbf{0.9422} $ & $ \textbf{0.0454} $ \\
    \multirow{2}*{leaves} & NeRF & $\textbf{21.586}$ & 0.7854 & 0.1629 \\
    ~ & ours & 21.487 & $ \textbf{0.7861} $ & $ \textbf{0.1599} $ \\
    \multirow{2}*{room} & NeRF & $ \textbf{33.852} $ & $ \textbf{0.9593} $  & $ \textbf{0.0721} $ \\
    ~ & ours & 33.760 & 0.9584 & 0.0735 \\
    \multirow{2}*{trex} & NeRF & $\textbf{30.287}$ & $\textbf{0.9501}$ & $\textbf{0.0601}$ \\
    ~ & ours & 30.222 & 0.9499 & 0.0605\\
  \bottomrule
\end{tabular}
\end{table}

\subsection{DoF Rendering}

\begin{figure*}[t]
  \centering
  \setlength{\abovecaptionskip}{2pt}
  \setlength{\belowcaptionskip}{2pt}
  \includegraphics[width=\linewidth]{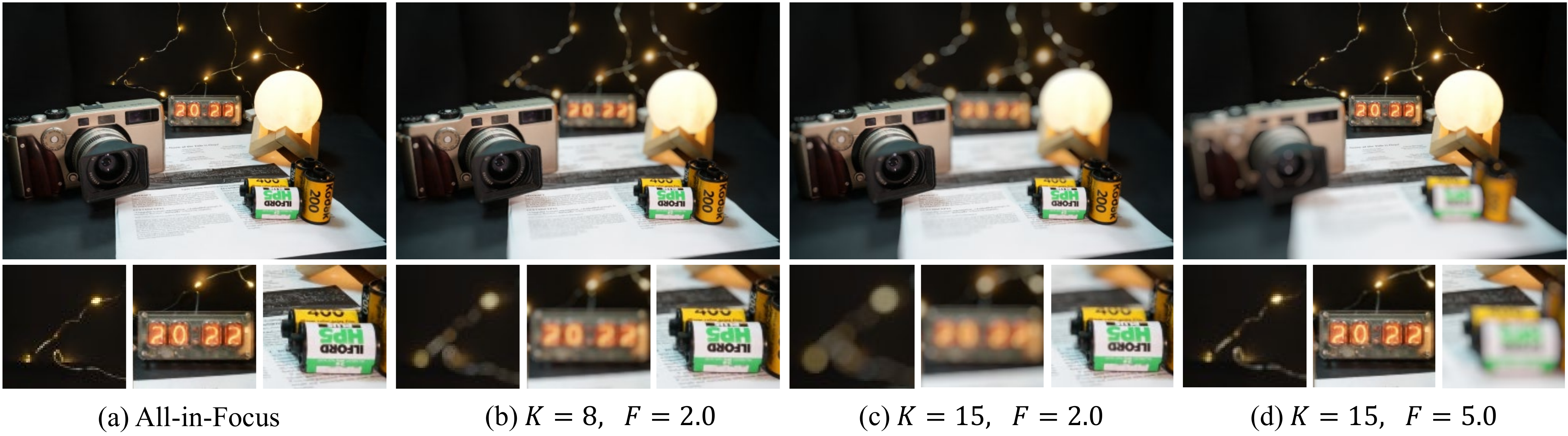} 
  \caption{ Visualization of adjustable DoF rendering. Apart from (a) all-in-focus rendering, we can manipulate DoF effect by changing aperture and focus settings. We set $F$ to $2.0$ and $K$ to (b) $8$ and (c) $15$ respectively to render images with various aperture parameters. By holding $K$ at $15$, we change $F$ to (d) $5.0$ to create images with different focus distance.}
  \Description{Rendering various views.}
  \label{fig:render-various}
\end{figure*}

Apart from representing all-in-focus 3D scenes with shallow DoF inputs, it is also possible to render the DoF effect with the optical modeling for volume rendering. 
By
changing the aperture 
or focus 
settings, we can manipulate the DoF effect in novel view synthesis. In Fig.~\ref{fig:render-various}, we visualize the effect of aperture parameter and focusing distance by using different rendering settings.

\begin{figure}[t]
  \centering
  \setlength{\abovecaptionskip}{2pt}
  \setlength{\belowcaptionskip}{2pt}
  \includegraphics[width=\linewidth]{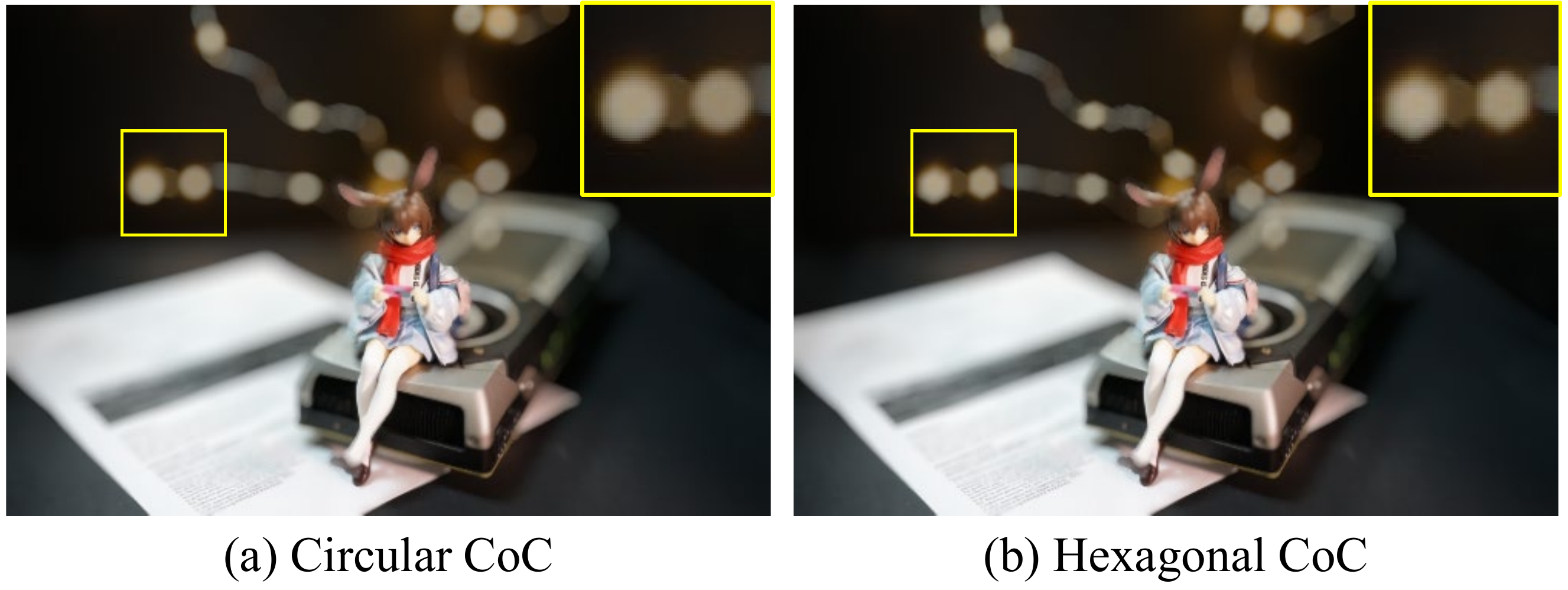}
  \caption{ Visualization of adjustable CoC shape rendering, such as (a) circular and (b) hexagonal. }
  \Description{Rendering various views.}
  \label{rendering}
\end{figure}

With different lens designs and configurations 
various CoC styles can be created. Our 
method can easily simulate this phenomenon by changing the shape of blur kernel. 
For example, we use a deformable polygonal kernel to create various DoF effects in Fig.~\ref{rendering}. 

\subsection{Parameter Analysis}

\begin{figure}[t]
  \centering
  \setlength{\abovecaptionskip}{2pt}
  \setlength{\belowcaptionskip}{2pt}
  \includegraphics[width=\linewidth]{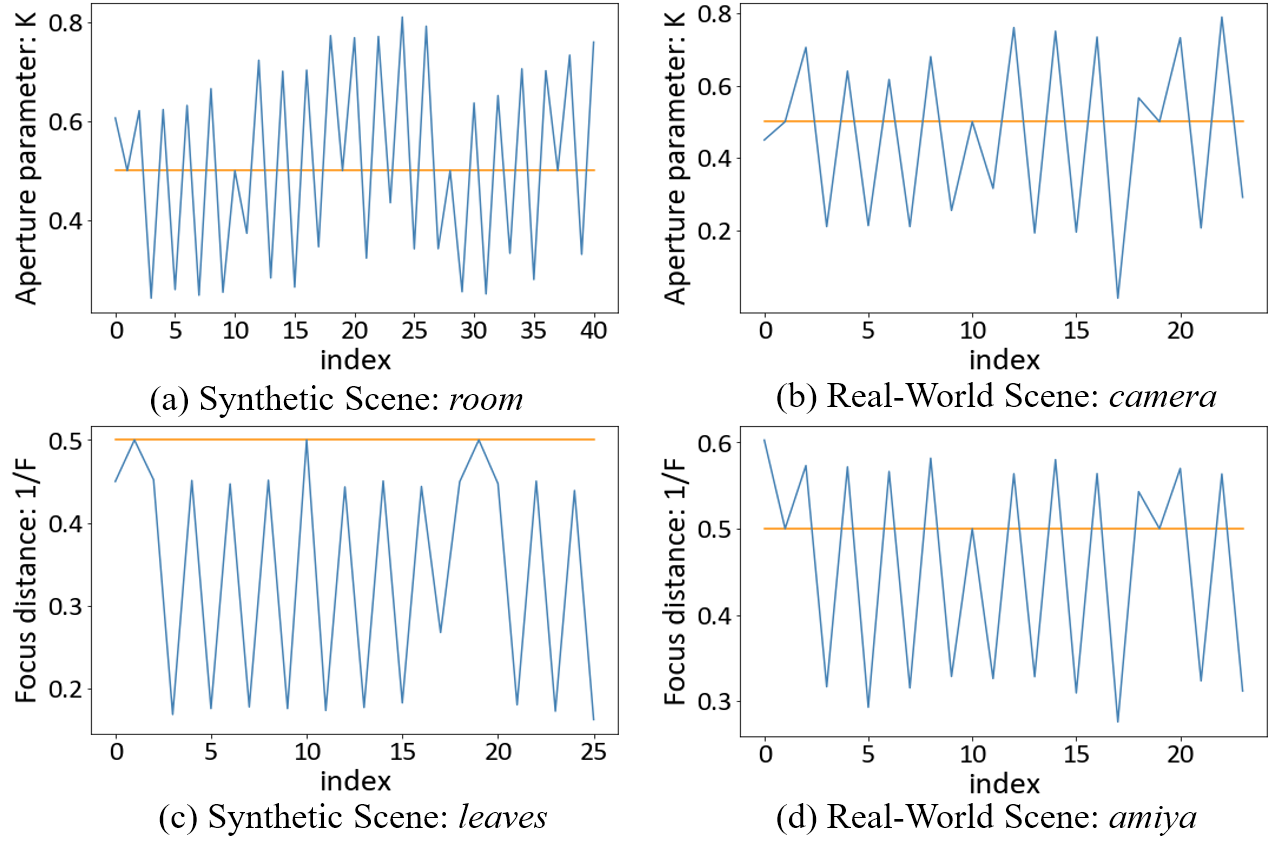}
  \caption{ Visualization of inferred aperture and focus parameters. 
  The orange lines 
  represent
  the initialized values, and we scale the aperture parameters by $10$.
  Some parameters remain unchanged because the corresponding views are held out for testing and not optimized during training.
  }
  \label{ablation}
\end{figure}

In this section, we 
construct 
image sets with mixing patterns to verify the validity of the estimated aperture parameters and focus distances, because accurate apertures and focus distances cannot be obtained in both the real-world dataset and synthetic dataset. Specifically, when validating the focus distance, we use foreground-focused images 
as 
even-indexed 
views, and background-focused images 
as 
odd-indexed 
views. When validating the aperture parameter, on odd-indexed 
views we use all-in-focus images, which can be seen as images taken using small aperture; on even-indexed 
views we use shallow DoF images, which can be seen as images taken using wide aperture. We visualize the value of apertures and focus distances of every image before and after optimization in Fig.~\ref{ablation}. The results show that the prediction of both aperture parameters and focus distances are optimized to the similar mixing pattern of the training set, which demonstrates that the estimated aperture parameters and focus distances are reasonable and valid.

\begin{table}[t]
  \setlength{\abovecaptionskip}{5pt}
  \setlength{\belowcaptionskip}{5pt}
  \caption{Ablation studies of components of our model. ``Aperture'' and ``Focus'' denote learnable aperture parameters and focus distances respectively.}
  \label{tab:ablation}
  \renewcommand\tabcolsep{3pt}
  \label{tab:exp-ablation}
  \begin{tabular}{ccccccc}
    \toprule
    Scene & Model & Aperture & Focus & PSNR$ (\uparrow) $ & SSIM$ (\uparrow) $ & LPIPS$ (\downarrow) $\\
    \midrule
    \multirow{4}*{camera} & NeRF &  --   &  --  & 25.742 & 0.8723 & 0.1657 \\
    ~ & ours & $ \checkmark $ & $ \times $  & 26.639 & 0.8989 & 0.1338 \\
    ~ & ours & $ \times $ & $ \checkmark $ & 25.855 & 0.8786 & 0.1532 \\
    ~ & ours & $ \checkmark $ & $ \checkmark $ & $\textbf{26.962}$ & $\textbf{0.9045}$ & $\textbf{0.1280}$ \\
  \bottomrule
\end{tabular}
\end{table}

To validate the design choice of our 
approach, we also conduct an ablation study 
on the scene \textit{camera}. 
From the results of different combinations in Table~\ref{tab:ablation}, one can observe that, 
i) explicit modeling of the aperture parameters and the focus distance benefits novel view synthesis, 
and ii) the aperture parameter and the focus distance can promote each other when they are jointly optimized.

\section{Conclusion}

In this work, we 
present DoF-NeRF, a novel 
framework for recovering sharp 3D scenes from sparse shallow DoF images. To achieve this, we 
model the CoC to simulate the radiance scattered between rays and introduce 
learnable parameters to enable direct manipulation of the DoF effect. Comprehensive experiments are conducted on both synthetic and real-world datasets, where DoF-NeRF not only performs comparably with NeRF in the all-in-focus setting, but also can synthesize all-in-focus novel views conditioned on shallow DoF inputs. Moreover, by changing the aperture parameter or the focus distance, DoF-NeRF 
can achieve controllable DoF 
rendering from novel viewpoints.

\begin{acks}
This work was funded by the DigiX Joint Innovation Center of Huawei-HUST.
\end{acks}

\bibliographystyle{ACM-Reference-Format}
  \bibliography{arxiv_submission}

\clearpage
\appendix

\section{Appendix Summary}
The appendix 
involves 
the following contents:
\begin{itemize}[leftmargin=*]
\item Table of notations (Table~\ref{tab:notations}).
\item Physical model of DoF.
\item Algorithm details of the concentrate-and-scatter method.
\item Implementation details.
\item Datasets details.
\item Full experimental reports and more qualitative results.
\item More synthesis of DoF effect. 
\end{itemize}

\begin{table}[h]
  \caption{Table of notations}\vspace{-10pt}
  \label{tab:notations}
  \begin{tabular}{cc}
    \toprule
    Notation & Description \\
    \midrule
    $N$ & Number of the images in the training set \\
    $N_s$ & Number of the sample points on a single ray \\
    $N_{patch}$ & Size of the ray patches \\
    $N_{anchor}$ & Distance between adjacent anchors\\
    $N_{iters}$ & Number of iterations for joint optimization\\
    $N_{pretrain}$ & Number of iterations for the first-stage optimization \\
    \midrule
    $\bm{x}$ & Spatial location \\
    $\bm{d}$ & Viewing direction \\
    $\bm{o}$ & Camera origin \\
    $\bm{p}$ & Pixel on the imaging plane \\
    $\bm{r}_{\bm{p}}$ & Ray specific to pixel $\bm{p}$ \\
    $\bm{c}$ & Radiance predicted by MLP \\
    \midrule
    $\alpha$ & Transparency predicted by MLP \\
    $f$ & Focal length \\
    $D$ & Aperture diameter \\
    $K$ & Aperture parameter \\
    $F$ & Focus distance \\
    $h$ & Depth of a spatial point along a ray \\
    $\Delta$ & Distance between adjacent sample points \\
    $h_c$ & Concentrated depth \\
    $\delta$ & Diameter of CoC \\
    $\delta_c$ & Diameter of CoC at the concentrated depth \\
    \midrule
    $\mathcal{I}$ & Set of input images \\
    $\mathcal{P}$ & Set of pixels \\
    $\mathcal{K}$ & Set of aperture parameters \\
    $\mathcal{F}$ & Set of focus distances \\
    \midrule
    $G_\Theta, \Theta$ & MLP network of NeRF and its parameters \\
    $\mathcal{L}$ & Loss \\
    $I$ & Input image \\
    $T$ & Accumulated transmittance \\
    $\hat{I}(\bm{p})$ & Predicted color of a pixel $\bm{p}$\\
    $\hat{I}_{ray}(\bm{p})$ & Diffused radiance of a pixel $\bm{p}$ \\
    $\hat{I}_{scatter}(\bm{p})$ & Scattering radiance from other rays \\
    $\hat{C}$ & Scattering radiance of a spatial point \\
    $\hat{C}_s(\bm{p})$ & Diffused radiance of a pixel $\bm{p}$ \\
    $K_{volume}$ & Volume rendering coefficient \\
  \bottomrule
\end{tabular}
\end{table}

\section{optics principle of DoF}

\begin{figure}[t]
  \centering
  \includegraphics[width=\linewidth]{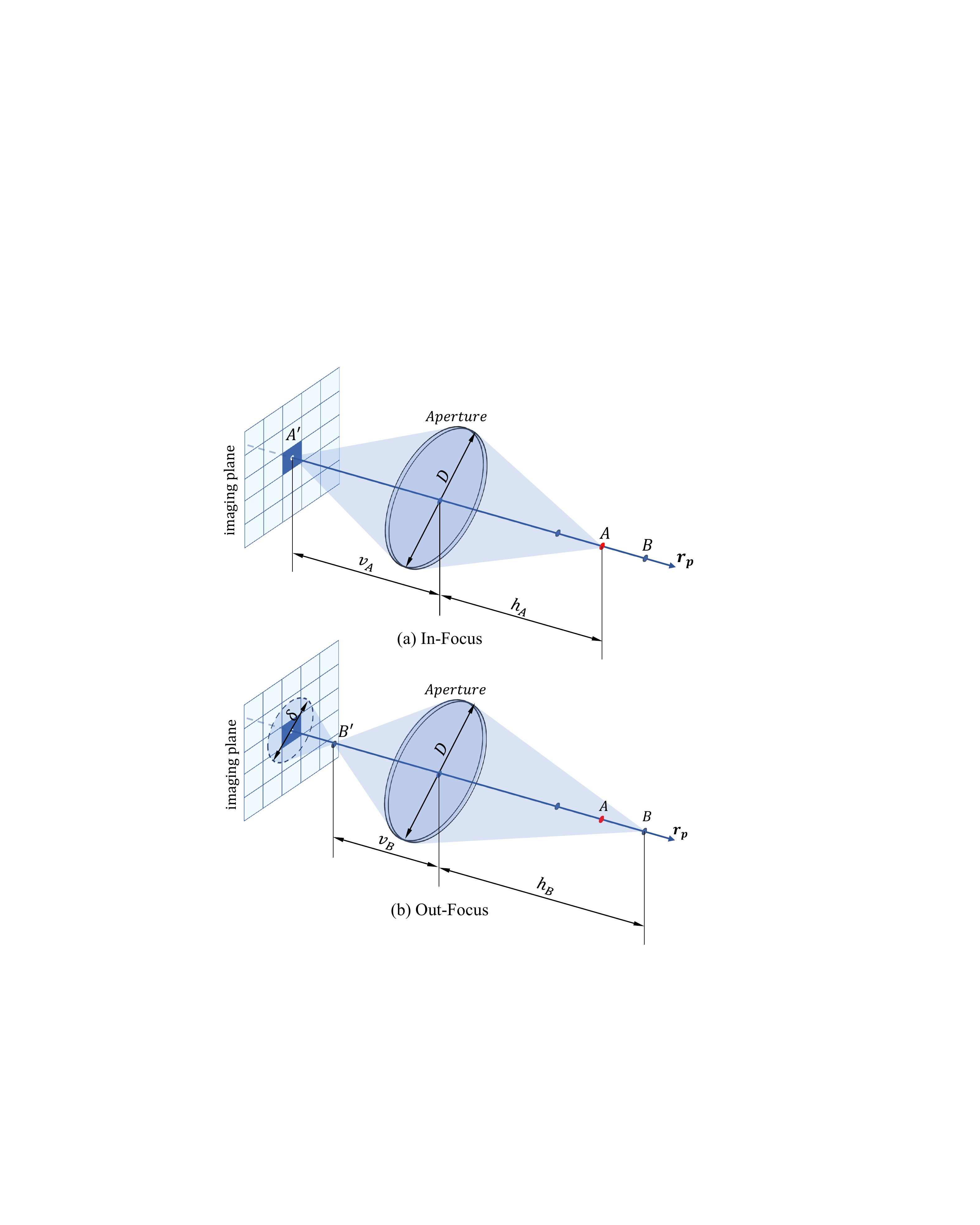}\vspace{-10pt}
  \caption{Principle of CoC and DoF.}
  \Description{Visualization of real-world dataset.}
  \label{optics}
\end{figure}

In this section, 
we provide a detailed derivation of CoC diameters. 
Given an ideal optical lens system with focal length $f$ and aperture diameter $D$, lights emitted from spatial points $A$ and $B$ on the ray $\bm{r}_{\bm{p}}$ with object distance $h_A$ and $h_B$ converge on points $A'$ and $B'$ with imaging distance $v_A$ and $v_B$, respectively (Fig.~\ref{optics}).
The relation between $h_A$ and $v_A$ (similarly, $h_B$ and $v_B$) follows the Gaussian formula
\begin{equation}\vspace{-5pt}
    \frac{1}{h_A}+\frac{1}{v_A}=\frac{1}{f}\,,
    \label{eq:gaussian1}
\end{equation}
\begin{equation}
    \frac{1}{h_B}+\frac{1}{v_B}=\frac{1}{f}\,.
    \label{eq:gaussian2}
\end{equation}

As shown in Fig~\ref{optics}, 
point $A$ is imaged to the converging point $A'$ on the imaging plane. In contrast, point $B$ is projected to a CoC with diameter $\delta_B$ on the imaging plane.
According to the properties of similar triangles, $\delta_B$ can be 
computed 
by 
\begin{equation}
    \frac{\delta_B}{D} = \frac{v_A-v_B}{v_B}\,.
\label{eq:sim_trian}
\end{equation}
Substituting 
Eq.\eqref{eq:gaussian1} and Eq.\eqref{eq:gaussian2} 
into 
Eq.\eqref{eq:sim_trian}
takes the form 
\begin{equation}
    \delta_B =  fD\times\frac{h_B-h_A}{h_B(h_A-f)}\,.
\end{equation}
As the lens system focuses on point $A$, we can 
replace 
$h_A$ 
with 
focus distance $F$: 
\begin{equation}
    \delta_B =  fD\times\frac{h_B-F}{h_B(F-f)}\,.
\end{equation}

For an out-focus point $C$ with object distance $h_C$ smaller than focus distance $F$, 
the diameter of CoC $\delta_C$ is formed as 
\begin{equation}
    \delta_C =  fD\times\frac{F-h_C}{h_C(F-f)}\,.
\end{equation}
Thus the diameter of CoC can be formulated by 
\begin{equation}
    \delta = fD\times\frac{\left|F-h_t\right|}{h_t(F-f)}\,,
    \label{eq:delta}
\end{equation}
where $h_t$ denotes the object distance of a spacial point.
Since the focus distance $ F $ and object distance $ h_t $ are often 
much
larger than the focal length $ f $, we can 
modify 
Eq.~\eqref{eq:delta} 
such that
\begin{equation}
    \delta(h_t) = fD\times\frac{\left| h_t-F\right|}{Fh_t}=fD\times\left|\frac{1}{F}-\frac{1}{h_t}\right| \,.
\end{equation}

\section{Algorithm Details}

Here 
we present the implementation details of the Concentrate-and-Scatter algorithm.

As mentioned in the main paper, the concentration of the radiance follows the original volume rendering. 
We implement the radiance scattering method by a pixel-wise rendering algorithm (see Algorithm~\ref{alg:1}). 
Assuming that the aperture shape is circular, 
we first compute the signed defocus map $S$ 
with 
concentrated depth $H_c$, aperture parameter $K$ and focus distance $F$. We apply a gamma transformation to transform radiance $C$ to linear space. Two accumulation buffers $W$ and $I$ are 
initialized with zero. 
Function \textit{TraversePatch}() is adopted to traverse all pixels of $C$. The scattering radius $r_i$ can be calculated by the absolute value of $S_i$. We then traverse the neighboring pixels of $p_i$ by the function \textit{TraverseNeighbor}(). We  calculate the weight $w_i$ from $p_i$ to its neighboring pixel $p_j$. For pixel $p_i$, its radiance can only scatter to $p_j$ if scattering radius $r_i$ is larger than the distance $l_{ij}$ between $p_i$ and $p_j$. 

To produce smooth and natural DoF effect, a soft CoC kernel in the calculation of weight $w_{ij}$ is adopted. We additionally divide $w_{ij}$ by the square of $r_i$ due to the uniform distribution of radiance. The calculated $w_{ij}$ and radiance $C_i$ weighted by $w_{ij}$ are then accumulated in $W_j$ and $I_j$, respectively. After traversing all pixels, the rendering result $B_{cr}$ can be obtained by the element-wise division of $I$ and $W$. 
A inverse gamma transformation is applied subsequently.

To create polygonal CoC, we can modify row 8 by multiplying a factor $k_{ij}$ to $r_i$. The factor $k_{ij}$ is defined by
\begin{equation}
    k_{ij}=\frac{sin\left(\frac{\pi}{2}-\frac{\pi}{n}\right)}{sin\left(\frac{\pi}{2}-\frac{\pi}{n}+mod\left(\left|arctan \left(\frac{l_{ij}^y}{l_{ij}^x}\right)+\phi\right|,\frac{2\pi}{n}\right)\right)}\,,
\end{equation}
where $n$ denotes the number of aperture blades and $\phi$ represents the rotation angle of polygonal CoC. $l_{ij}^x$ and $l_{ij}^y$ denote the horizontal and vertical component of distance $l_{ij}$. 

\begin{algorithm}[t]
    \small
    \caption{Pixel-wise scattering method}
    \label{alg:1}
    
    
    \KwIn{Concentrated radiance $C$, concentrated depth $H_c$, aperture parameter $K$, focus distance $F$, gamma value $\gamma$}
    \KwOut{Scattering result $B_{cr}$}
    $S \leftarrow K\cdot(\frac{1}{H_c} - \frac{1}{F})$\,\;
    $C \leftarrow (C)^\gamma$\,\;
    $W \leftarrow [0]$\,\;
    $I \leftarrow [0]$\,\;
    \For {$p_i \leftarrow$ \textit{TraversePatch} ($C$)}
    {
        $r_i \leftarrow \lvert S_i \rvert$\,\;
        \For {$p_j \leftarrow$ \textit{TraverseNeighbor} ($p_i,r_i$)}
        {
            $w_{ij} \leftarrow \frac{0.5+0.5\tanh{(4(r_i-l_{ij}))}}{{r_i}^2+0.2}$\,\;
            $W_j \leftarrow W_j + w_{ij}$\,\;
            $I_j \leftarrow I_j + w_{ij}\cdot C_i$\,\;
        }
    }
    $B_{cr} \leftarrow \dfrac{I}{W}$\,\;
    $B_{cr} \leftarrow (B_{cr})^{\frac{1}{\gamma}}$\,\;
\end{algorithm}

\section{Implementation Details}
We use a batch size of 1024 rays for NeRF and DS-NeRF. 
The learning rate of NeRF and DS-NeRF is set to $0.0005$ and decays exponentially to one-tenth for every $250000$ steps. 
For NeRF and DS-NeRF, we use 64 samples for the coarse network and 64 samples for the fine network. As explained in the main paper, we adopt a two-stage optimization: each stage takes $200k$ iterations, where $N_{pretrained}$ and $N_{iters}$ are set to $200k$ and $400k$, respectively. For all-in-focus inputs, we change $N_{pretrained}$ and $N_{iters}$ to $400k$ and $800k$, respectively. At the second optimization stage, we set $N_{patch}$ to 48 and $N_{anchor}$ to 16.
The aperture and focus parameters are both initialized with 0.5 for DoF-NeRF. 
For ablation study, the aperture and focus parameters are set to 0.5 for initialization and fixed according to the experiment settings. 

\section{Dataset Details}

\subsection{Real-World Dataset}
\begin{figure}[t]
  \centering
  \includegraphics[width=\linewidth]{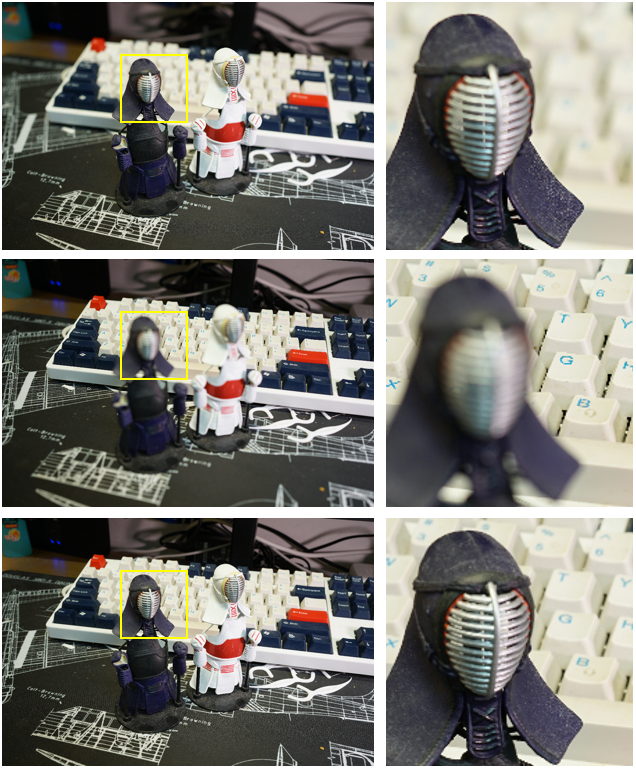}
  \caption{ Visualization of the scene \textit{kendo} in the real-world dataset. Each row respectively represents the foreground-focusing shallow DoF image, background-focusing shallow DoF image, and wide DoF image. We zoom in the yellow boxes in the images and present them on the right.
  }
  \Description{Visualization of real-world dataset.}
  \label{kendo_vis}
\end{figure}

The real-world dataset consists of 7 scenes: \textit{amiya}, \textit{camera}, \textit{plant}, \textit{kendo}, \textit{desk}, \textit{shelf}, and \textit{turtle}. Each scene contains $20\sim30$ image triplets. Each triplet includes a wide DoF image taken with small aperture and two images taken with large aperture focusing on the foreground and background respectively (Fig.~\ref{kendo_vis}). 

\begin{figure}[t]
  \centering
  \includegraphics[width=\linewidth]{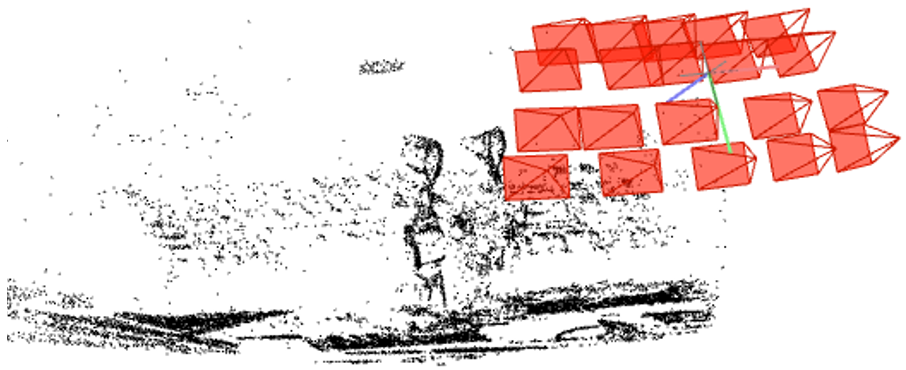}
  \caption{ Visualization of the predicted camera poses and reconstructed sparse point cloud by COLMAP~\cite{schonberger2016structure}.}
  \Description{Visualization of colmap camera poses.}
  \label{colmap}
\end{figure}

\begin{figure}[t]
  \centering
  \includegraphics[width=\linewidth]{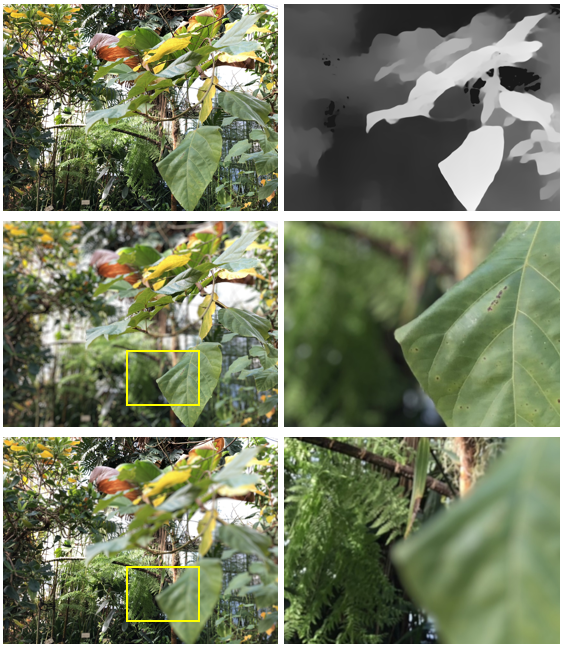}
  \caption{ Visualization of the scene \textit{leaves} in the synthetic dataset. The first row shows the original image from the Real Forward-Facing dataset~\cite{mildenhall2020nerf} and the predicted disparity map from DPT~\cite{Ranftl_2021_ICCV}. The second and third rows are the shallow DoF images rendered by BokehMe~\cite{Peng2022BokehMe}. 
  We zoom in the yellow boxes in the images and present them on the right.
  }
  \Description{Visualization of synthetic dataset.}
  \label{leaves_vis}
\end{figure}

All images are taken by a Sony ILCE-7RM2 camera with an FE 35mm f/1.8 lens. We secure the camera to a tripod and use remote control when taking images, in order to preserve identical camera parameters. Instead of using the maximum aperture to create shallow DoF, we use the aperture of $f/4$ to avoid distinct lens distortion and vignette. The camera parameters are generated by COLMAP~\cite{schonberger2016structure}.
As an example, we visualize the predicted camera poses and sparse point cloud of the scene \textit{kendo} in Fig.~\ref{colmap}. 
The original resolution of the images in the real-world dataset is $3976\times 2652$. 
We set the resolution to $497\times 331$ for training and evaluation.

\subsection{Synthetic Dataset}
The synthetic dataset is generated from the Real Forward-Facing dataset~\cite{mildenhall2020nerf} which consists of 8 scenes: \textit{fern}, \textit{flower}, \textit{fortress}, \textit{horns}, \textit{leaves}, \textit{orchids}, \textit{room}, and \textit{trex}. Since the images are captured with a handheld cellphone, the original images from the Real Forward-Facing dataset are used as wide DoF images in the triplets. The shallow DoF images are generated based on depth estimation and a recent single-image bokeh rendering framework. For each wide DoF image, we use DPT~\cite{Ranftl_2021_ICCV} to generate the disparity map and  BokehMe~\cite{Peng2022BokehMe} to synthesize the shallow DoF images. The blur parameter of BokehMe is set to 20 and the focus distances are set to 0.1 and 0.9 to simulate background focused and foreground focused shallow DoF images (Fig.~\ref{leaves_vis}). The original resolution of the images in the synthetic dataset is $4032\times 3024$. We set the resolution to $504\times 378$ for training and evaluation.

\section{Experimental Results}
In this section, we present full reports of experiments conducted in main paper and more qualitative comparisons. 

\begin{table}[t]
  \caption{Comparison of NeRF~\cite{mildenhall2020nerf} and our framework in the real-world dataset.}
  \label{tab:exp-rw-full}
  \begin{tabular}{ccccc}
    \toprule
    Scene & Model & PSNR$ (\uparrow) $ & SSIM$ (\uparrow) $ & LPIPS$ (\downarrow) $\\
    \midrule
    \multirow{2}*{amiya} & NeRF & 26.924 & 0.9092 & 0.1633 \\
    ~ & ours & $ \textbf{28.311} $ & $ \textbf{0.9289} $ & $ \textbf{0.1370} $ \\
    \multirow{2}*{camera} & NeRF & 25.593 & 0.8862 & 0.1574 \\
    ~ & ours & $ \textbf{27.714} $ & $ \textbf{0.9134} $ & $ \textbf{0.1259} $ \\
    \multirow{2}*{plant} & NeRF & 28.272 & 0.8961 & 0.1581 \\
    ~ & ours & $ \textbf{30.317} $ & $ \textbf{0.9290} $ & $ \textbf{0.1178} $\\
    \multirow{2}*{turtle} & NeRF & 33.531 & 0.9566 & 0.0939  \\
    ~ & ours & $ \textbf{34.965} $ & $ \textbf{0.9647} $ & $ \textbf{0.0823} $\\
    \multirow{2}*{kendo} & NeRF & 18.457 & 0.6643 & 0.2906 \\
    ~ & ours & $ \textbf{19.311} $ & $ \textbf{0.6970} $ & $ \textbf{0.2773} $\\
    \multirow{2}*{desk} & NeRF & 29.618 & 0.9374 & 0.1099 \\
    ~ & ours & $ \textbf{30.548} $ & $ \textbf{0.9426} $ & $ \textbf{0.1047} $\\
    \multirow{2}*{shelf} & NeRF & 31.552 & 0.9515 & 0.0811 \\
    ~ & ours & $ \textbf{32.002} $ & $ \textbf{0.9565} $ & $ \textbf{0.0723} $\\
  \bottomrule
\end{tabular}
\end{table}

\begin{figure*}[t]
  \centering
  \setlength{\abovecaptionskip}{2pt}
  \setlength{\belowcaptionskip}{2pt}
  \includegraphics[width=\linewidth]{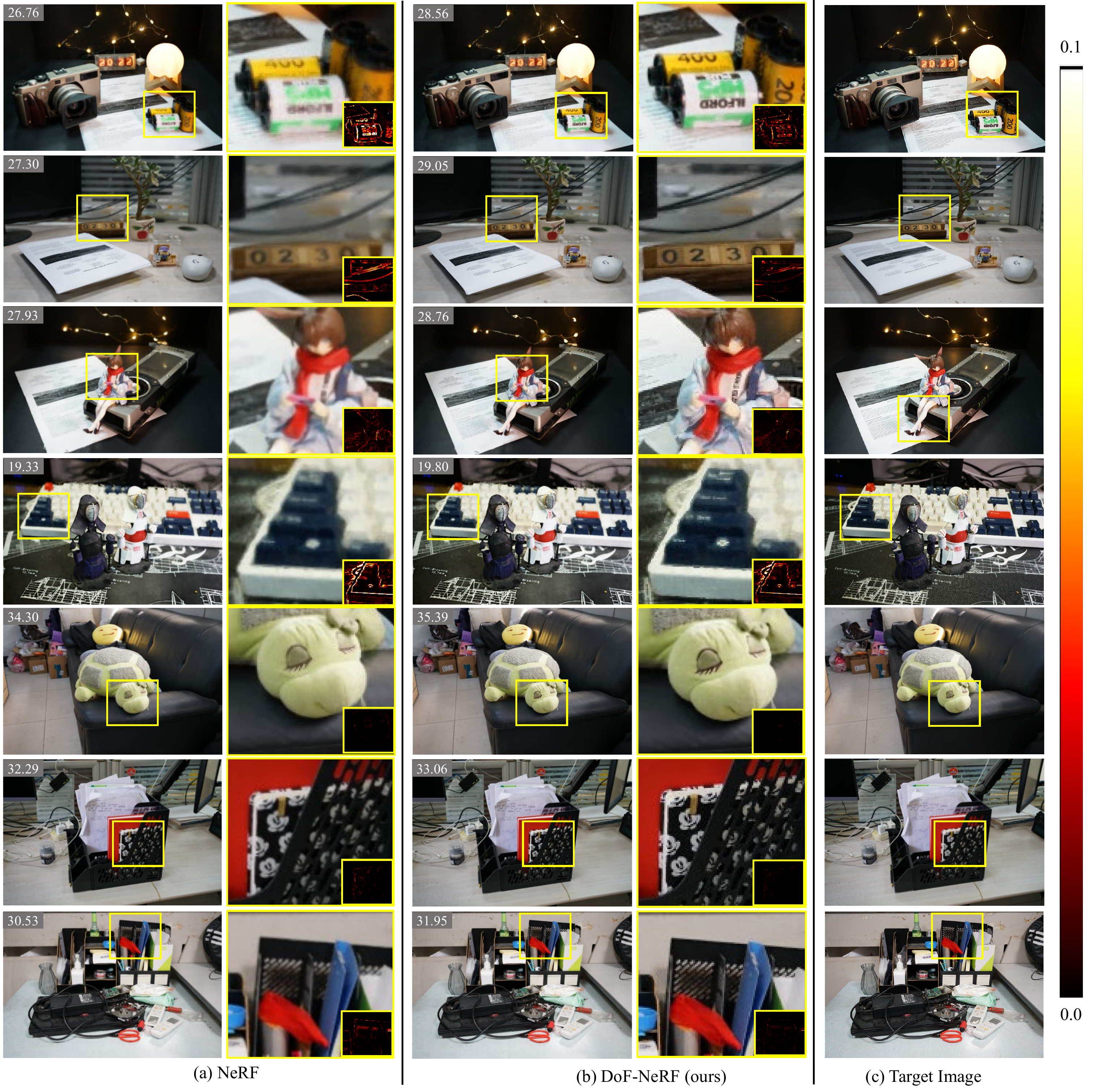}
  \caption{ Additional comparison of NeRF~\cite{mildenhall2020nerf} and our approach 
  with shallow DoF inputs on the real-world dataset. The first and third column show the images rendered by NeRF and our approach, respectively. PSNR is shown at the upper left corner. 
  We zoom in all the yellow boxes 
  and present them 
  in the second and fourth column with error maps ($0$ to $0.1$ pixel intensity range) shown at the lower right corner. }
  \label{rw-full}
\end{figure*}

\begin{figure*}[t]
  \centering
  \setlength{\abovecaptionskip}{2pt}
  \setlength{\belowcaptionskip}{2pt}
  \includegraphics[width=0.90\linewidth]{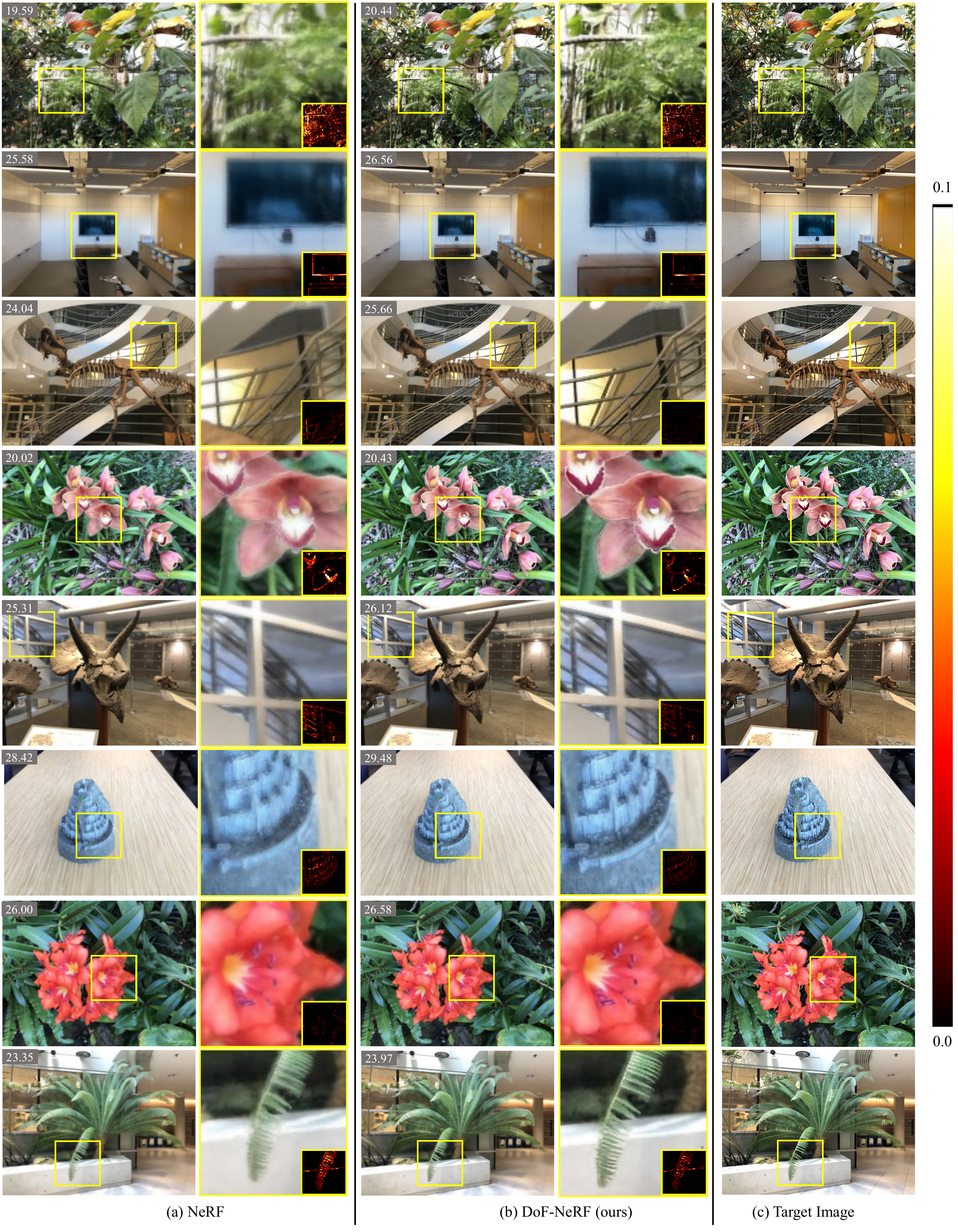}
  \caption{ Additional comparison of NeRF~\cite{mildenhall2020nerf} and our approach 
  with shallow DoF inputs on the synthetic dataset. The first and third column show the images rendered by NeRF and our approach, respectively. PSNR is shown at the upper left corner. 
  We zoom in all the yellow boxes 
  and present them 
  in the second and fourth column with error maps ($0$ to $0.1$ pixel intensity range) shown at the lower right corner. }
  \label{syn-full}
\end{figure*}

We extend Table~\ref{tab:exp-syn} and Table~\ref{tab:exp-rw} in the main paper by conducting experiments in other scenes of the synthetic dataset and real-world dataset. Table~\ref{tab:exp-rw-full} and Table~\ref{tab:exp-syn-full} report the rendering quality of vanilla NeRF and our model in the real-world dataset and synthetic dataset, respectively. Fig.~\ref{rw-full} and Fig.~\ref{syn-full} respectively show 
additional 
qualitative results using the real-world and synthetic shallow DoF inputs. 
Given 
shallow DoF inputs, 
our method demonstrates better perceptual qualities than vanilla NeRF when rendering all-in-focus novel views on both the real-world dataset and synthetic dataset. 

\begin{table}[t]
  \caption{Comparison of NeRF~\cite{mildenhall2020nerf} and our framework in the synthetic dataset.}
  \label{tab:exp-syn-full}
  \begin{tabular}{ccccc}
    \toprule
    Scene & Model & PSNR $ (\uparrow) $ & SSIM$ (\uparrow) $ & LPIPS$ (\downarrow) $\\
    \midrule
    \multirow{2}*{fortress} & NeRF & 28.142 & 0.7826 & 0.2011\\
    ~ & ours & $ \textbf{29.168} $ & $ \textbf{0.8099} $ & $ \textbf{0.1830} $ \\
    \multirow{2}*{leaves} & NeRF & 19.450 & 0.6541 & 0.3190 \\
    ~ & ours & $ \textbf{20.025} $ & $ \textbf{0.7000} $ & $ \textbf{0.2766} $ \\
    \multirow{2}*{room} & NeRF & 26.668 & 0.8743 & 0.1961 \\
    ~ & ours & $ \textbf{29.443} $ & $ \textbf{0.9135} $  & $ \textbf{0.1502} $ \\
    \multirow{2}*{trex} & NeRF & 24.433 & 0.8379 & 0.1723 \\
    ~ & ours & $\textbf{25.726}$ & $\textbf{0.8744}$ & $\textbf{0.1564}$\\
    \multirow{2}*{fern} & NeRF & 23.202 & 0.7232 & 0.3125 \\
    ~ & ours & $\textbf{23.407}$ & $\textbf{0.7311}$ & $\textbf{0.3055}$\\
    \multirow{2}*{flower} & NeRF & 26.339 & 0.8209 & 0.1999 \\
    ~ & ours & $\textbf{27.075}$ & $\textbf{0.8474}$ & $\textbf{0.1678}$\\
    \multirow{2}*{horns} & NeRF & 24.260 & 0.7554 & 0.2983 \\
    ~ & ours & $\textbf{24.829}$ & $\textbf{0.7766}$ & $\textbf{0.2759}$\\
    \multirow{2}*{orchids} & NeRF & 19.755 & 0.6401 & 0.2932 \\
    ~ & ours & $\textbf{20.048}$ & $\textbf{0.6678}$ & $\textbf{0.2718}$\\
  \bottomrule
\end{tabular}
\end{table}

\begin{table}[t]
  \caption{Comparison of DSNeRF~\cite{deng2021depth} and our framework in the synthetic dataset.}
  \label{tab:exp-dsnerf-all}
  \begin{tabular}{ccccc}
    \toprule
    Scene & Model & PSNR$ (\uparrow) $ & SSIM$ (\uparrow) $ & LPIPS$ (\downarrow) $\\
    \midrule
    \multirow{2}*{fortress} & DSNeRF & 28.493 & 0.7609 & 0.2373\\
    ~ & ours & $ \textbf{29.704} $ & $ \textbf{0.8112} $ & $ \textbf{0.1970} $ \\
    \multirow{2}*{leaves} & DSNeRF & 17.872 & 0.5096 & 0.4358 \\
    ~ & ours & $ \textbf{18.312} $ & $ \textbf{0.5689} $ & $ \textbf{0.3840} $ \\
    \multirow{2}*{room} & DSNeRF & 26.8301 & 0.8641 & 0.2287 \\
    ~ & ours & $ \textbf{29.550} $ & $ \textbf{0.9045} $  & $ \textbf{0.1817} $ \\
    \multirow{2}*{trex} & DSNeRF & 22.334 & 0.7063 & 0.3438 \\
    ~ & ours & $\textbf{23.600}$ & $\textbf{0.7764}$ & $\textbf{0.2773}$\\
    \multirow{2}*{fern} & NeRF & 23.100 & 0.7050 & 0.3355 \\
    ~ & ours & $\textbf{23.640}$ & $\textbf{0.7307}$ & $\textbf{0.3116}$\\
    \multirow{2}*{flower} & NeRF & 25.388 & 0.7683 & 0.2517 \\
    ~ & ours & $\textbf{26.392}$ & $\textbf{0.8131}$ & $\textbf{0.2024}$\\
    \multirow{2}*{horns} & NeRF & 21.562 & 0.5938 & 0.4625 \\
    ~ & ours & $\textbf{22.487}$ & $\textbf{0.6491}$ & $\textbf{0.4111}$\\
    \multirow{2}*{orchids} & NeRF & 19.435 & 0.6017 & 0.3247 \\
    ~ & ours & $\textbf{19.872}$ & $\textbf{0.6500}$ & $\textbf{0.2876}$\\
  \bottomrule
\end{tabular}
\end{table}

\begin{figure*}[t]
  \centering
  \setlength{\abovecaptionskip}{2pt}
  \setlength{\belowcaptionskip}{2pt}
  \includegraphics[width=\linewidth]{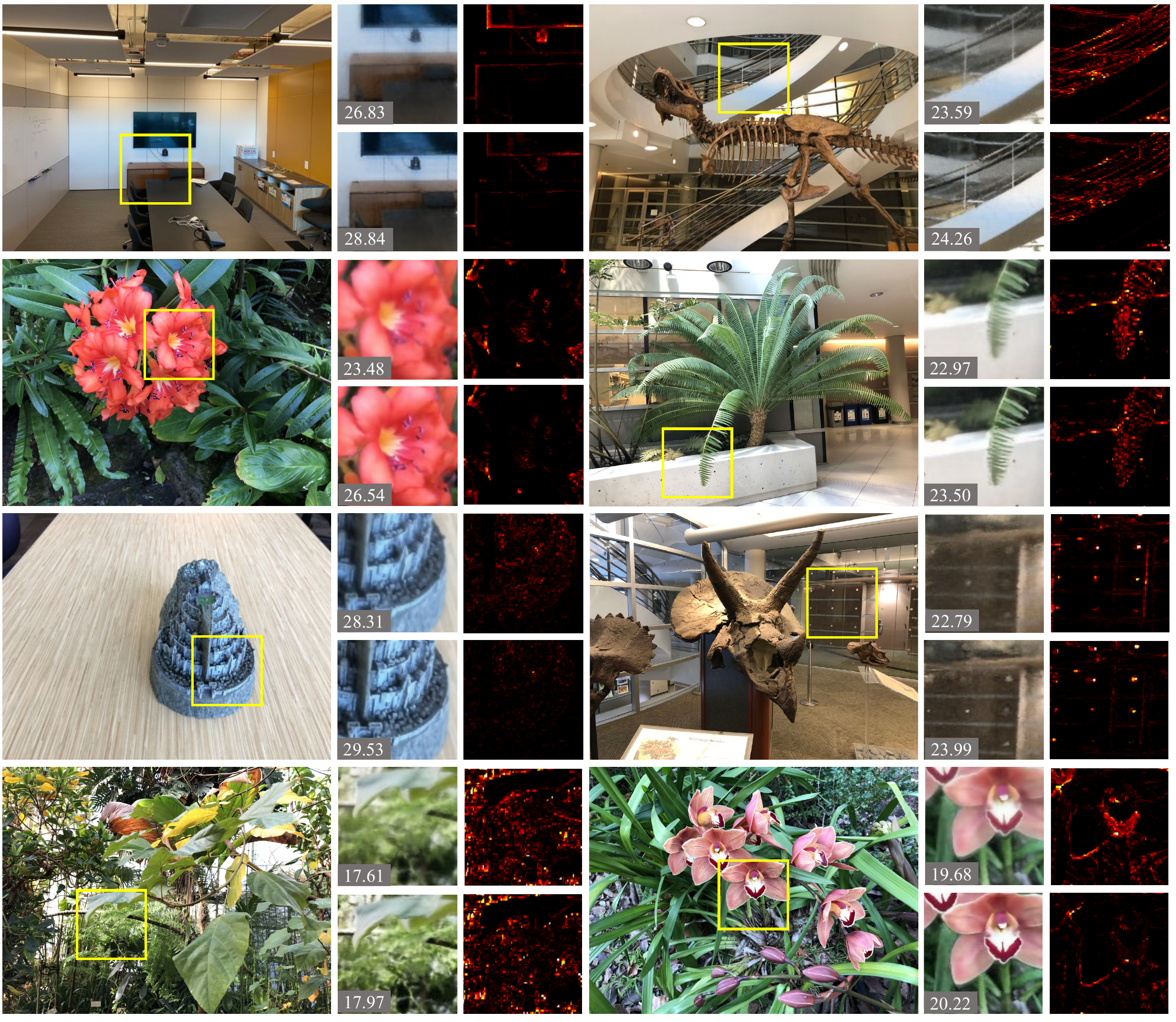}
  \caption{ 
    Additional comparison of DS-NeRF~\cite{deng2021depth} and our approach in the synthetic dataset. For each scene, the zoom-in of rendered images and error maps ($0$ to $0.2$ pixel intensity range) are presented. They are respectively obtained from DS-NeRF (first row) and our model (second row). For each subfigure, PSNR is shown on the lower left. 
  }
  \label{dsnerf-full}
\end{figure*}

Table~\ref{tab:exp-dsnerf-all} shows the rendering quality of our method using DS-NeRF as baseline on all scenes from the synthetic dataset. Our method can work as a plug-and-play module to improve the performance of NeRF-based method with shallow DoF inputs. 
Qualitative results are visualized in Fig.~\ref{dsnerf-full}.

Although our framework is originally designed for shallow DoF inputs, 
Table~\ref{tab:exp-aif-all} 
indicates that it shows comparable performance against vanilla NeRF using all-in-focus images as inputs. 

\begin{table}[t]
  \caption{Comparison of NeRF~\cite{mildenhall2020nerf} and our framework in the all-in-focus dataset.}
  \label{tab:exp-aif-all}
  \begin{tabular}{ccccc}
    \toprule
    Scene & Model & PSNR$ (\uparrow) $ & SSIM$ (\uparrow) $ & LPIPS$ (\downarrow) $\\
    \midrule
    \multirow{2}*{fortress} & NeRF & 33.973 & 0.9411 & 0.0469\\
    ~ & ours & $ \textbf{34.007} $ & $ \textbf{0.9422} $ & $ \textbf{0.0454} $ \\
    \multirow{2}*{leaves} & NeRF & $\textbf{21.5860}$ & 0.7854 & 0.1629 \\
    ~ & ours & 21.4869 & $ \textbf{0.7861} $ & $ \textbf{0.1599} $ \\
    \multirow{2}*{room} & NeRF & $ \textbf{33.852} $ & $ \textbf{0.9593} $  & $ \textbf{0.0721} $ \\
    ~ & ours & 33.760 & 0.9584 & 0.0735 \\
    \multirow{2}*{trex} & NeRF & $\textbf{30.287}$ & $\textbf{0.9501}$ & $\textbf{0.0601}$ \\
    ~ & ours & 30.222 & 0.9499 & 0.0605\\
    \multirow{2}*{fern} & NeRF & $\textbf{26.582}$ & 0.8490 & 0.1480 \\
    ~ & ours & 26.565 & $\textbf{0.8497}$ & $\textbf{0.1470}$\\
    \multirow{2}*{flower} & NeRF & $\textbf{30.135}$ & 0.9221 & 0.0643 \\
    ~ & ours & 30.093 & $\textbf{0.9224}$ & $\textbf{0.0626}$\\
    \multirow{2}*{horns} & NeRF & $\textbf{28.205}$ & 0.8879 & 0.1332 \\
    ~ & ours & 28.165 & $\textbf{0.8888}$ & $\textbf{0.1313}$\\
    \multirow{2}*{orchids} & NeRF & $\textbf{20.831}$ & 0.7362 & 0.1896 \\
    ~ & ours & 20.795 & $\textbf{0.7374}$ & $\textbf{0.1876}$\\
  \bottomrule
\end{tabular}
\end{table}

Table~\ref{tab:ablation_supp} shows the results of the ablation study conducted on scene \textit{room} from the synthetic dataset and on the scene \textit{camera} from the real-world dataset.

\begin{table}[t]
  \caption{Ablation studies of components of our model. ``Aperture'' and ``Focus'' denote learnable aperture parameters and focus distances respectively.}
  \label{tab:ablation_supp}
  \renewcommand\tabcolsep{3pt}
  \label{tab:exp-ablation-all}
  \begin{tabular}{ccccccc}
    \toprule
    Scene & Model & Aperture & Focus & PSNR$ (\uparrow) $ & SSIM$ (\uparrow) $ & LPIPS$ (\downarrow) $\\
    \midrule
    \multirow{4}*{room} & NeRF &  --   &  --  & 27.279 & 0.8863 & 0.1803 \\
    ~ & ours & $ \checkmark $ & $ \times $  & 27.799 & 0.8992 & 0.1640 \\
    ~ & ours & $ \times $ & $ \checkmark $ & 28.070 & 0.9032 & 0.1671 \\
    ~ & ours & $ \checkmark $ & $ \checkmark $ & $\textbf{28.361}$ & $ \textbf{0.9076} $ & $\textbf{0.1556}$ \\
    \midrule
    \multirow{4}*{camera} & NeRF &  --   &  --  & 25.742 & 0.8723 & 0.1657 \\
    ~ & ours & $ \checkmark $ & $ \times $  & 26.639 & 0.8989 & 0.1338 \\
    ~ & ours & $ \times $ & $ \checkmark $ & 25.855 & 0.8786 & 0.1532 \\
    ~ & ours & $ \checkmark $ & $ \checkmark $ & $\textbf{26.962}$ & $\textbf{0.9045}$ & $\textbf{0.1280}$ \\
  \bottomrule
\end{tabular}
\end{table}

\section{DoF Rendering}

\begin{figure*}[t]
  \centering
  \setlength{\abovecaptionskip}{2pt}
  \setlength{\belowcaptionskip}{2pt}
  \includegraphics[width=\linewidth]{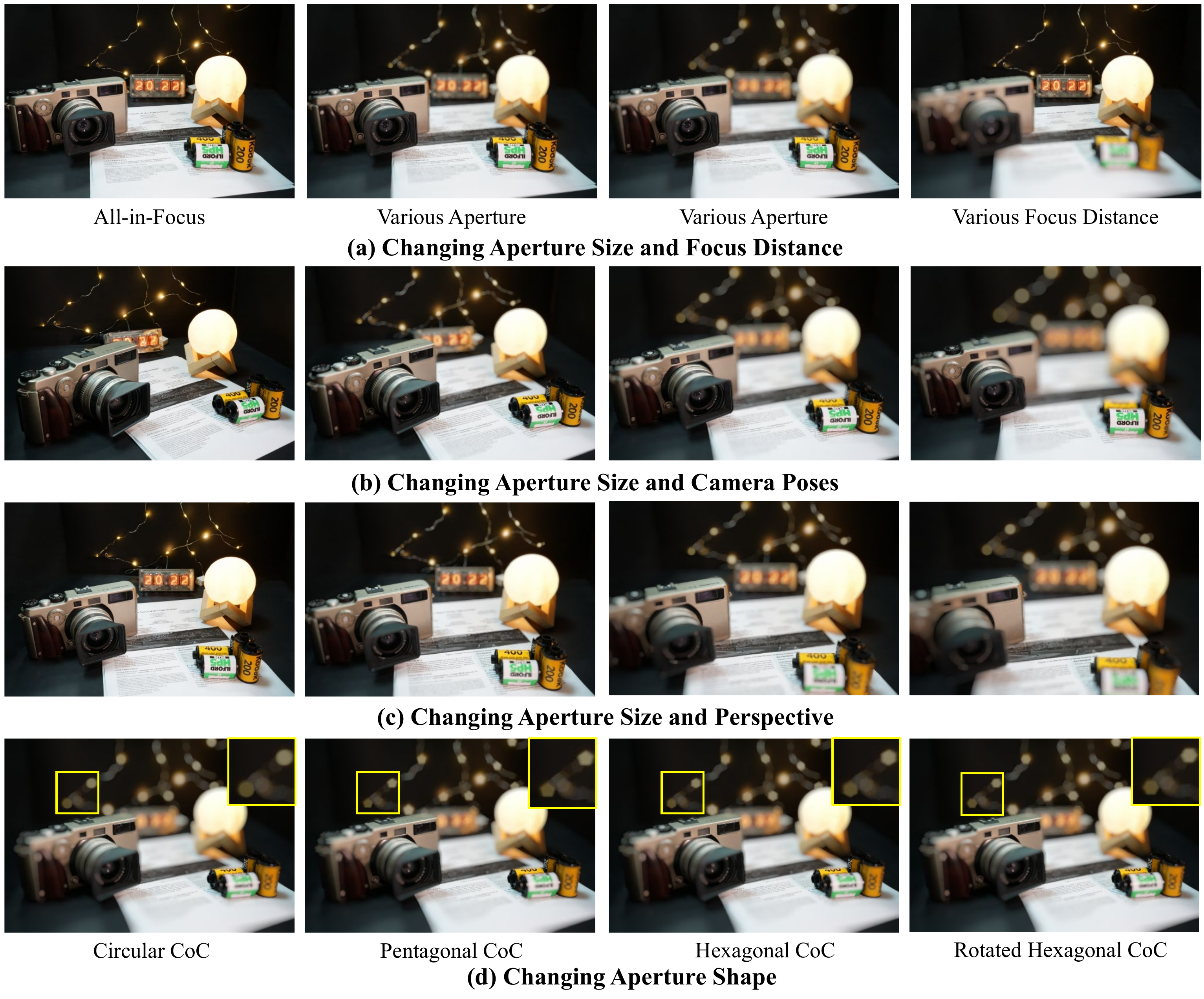}
  \caption{ More synthesis of controllable DoF effect. DoF-NeRF can manipulate the DoF rendering by changing (a) aperture size and focus distance, (b) camera poses, (c) perspective, and (d) aperture shape. }
  \label{render-bokeh}
\end{figure*}

In this section, we present more synthesis of controllable DoF rendering (Fig.~\ref{render-bokeh}). Benefiting from the explicit modeling of DoF effect, DoF-NeRF enables direct manipulation of DoF effect by adjusting virtual aperture size and focus distance. Compared with single-image DoF rendering method, DoF-NeRF is capable of rendering DoF effect from novel viewpoints by adjusting the corresponding camera parameters. 
With different lens designs and configurations 
such as 
the shape of aperture, various CoC styles can be created. 
Our concentrate-and-scatter method can easily simulate this phenomenon by changing the shape of blur kernel. 
For example, we use a deformable polygonal kernel to create various DoF effects.

\section{Cost of Training and Testing}
We compare the training and testing time of DoF-NeRF and vanilla NeRF on the same machine using a single NVIDIA RTX 3090 GPU. The following report are measured on the Scene Camera from the real-world dataset.

\subsection{Training}
We train each model for a total of $400k$ iterations, \textit{i.e.}, $400k$ iterations for vanilla NeRF, $200k$ iterations each for the first and second stage of DoF-NeRF. The batch size is set to $1024$ rays for NeRF and the first stage of DoF-NeRF. We set $N_{patch}$ to $32$ in order to calculate the same amount of rays in every iteration. 
\begin{itemize}[leftmargin=*]
    \item Vanilla NeRF takes $6h26min$.
    \item DoF-NeRF takes $6h41min$, $3.8\%$ longer than vanilla NeRF.
\end{itemize}
Both models take about 6GB GPU memory during training. 

\subsection{Testing}
We render 120 images from different viewpoints using NeRF and DoF-NeRF model. All the synthesized images are of $497\times 331$ resolution. 
\begin{itemize}[leftmargin=*]
    \item Vanilla NeRF takes $403s$, \textit{i.e.}, $3.36s$ per image.
    \item DoF-NeRF takes $405s$, \textit{i.e.}, $3.38s$ per image, $0.60\%$ longer than vanilla NeRF.
\end{itemize}

\end{document}